\documentclass[10pt,twocolumn,letterpaper]{article}

\usepackage[pagenumbers]{format_cvpr2024/cvpr} 

\usepackage[pagebackref,breaklinks,colorlinks]{hyperref}

\usepackage[dvipsnames]{xcolor}

\usepackage{multirow}
\usepackage{pifont}
\usepackage{amsfonts}
\usepackage{bm}

\newcommand{\R}{\mathbb{R}}

\definecolor{duke-blue}{RGB}{0,83,155}
\definecolor{duke-navy}{RGB}{1,33,105}
\newcommand{\dblue}[1]{{\color{duke-blue}#1}}

\newcommand{\cmark}{\ding{52}}

\makeatletter
\def\thickhline{%
  \noalign{\ifnum0=`}\fi\hrule \@height \thickarrayrulewidth \futurelet
   \reserved@a\@xthickhline}
\def\@xthickhline{\ifx\reserved@a\thickhline
               \vskip\doublerulesep
               \vskip-\thickarrayrulewidth
             \fi
      \ifnum0=`{\fi}}
\makeatother

\newlength{\thickarrayrulewidth}
\usepackage[toc,page,header]{appendix}
\usepackage{minitoc}
\usepackage{times}
\usepackage{epsfig}
\usepackage{graphicx}
\usepackage{amsmath}
\usepackage{amssymb}

\usepackage[capitalize]{cleveref}
\crefname{section}{Sec.}{Secs.}
\Crefname{section}{Section}{Sections}
\crefname{table}{Tab.}{Tabs.}
\Crefname{table}{Table}{Tables}

\title{\textbf{\dblue{UFD-PRiME}: \dblue{U}nsupervised Joint Learning of Optical \dblue{F}low and Stereo \dblue{D}epth through \dblue{P}ixel-Level \dblue{Ri}gid \dblue{M}otion \dblue{E}stimation}}
\author{Shuai Yuan and Carlo Tomasi\\
Duke University, Durham NC, USA
}
\date{}

\begin{document}

\maketitle


\begin{abstract}
Both optical flow and stereo disparities are image matches and can therefore benefit from joint training. Depth and 3D motion provide geometric rather than photometric information and can further improve optical flow. Accordingly, we design a first network that estimates flow and disparity jointly and is trained without supervision. A second network, trained with optical flow from the first as pseudo-labels, takes disparities from the first network, estimates 3D rigid motion at every pixel, and reconstructs optical flow again. A final stage fuses the outputs from the two networks. In contrast with previous methods that only consider camera motion, our method also estimates the rigid motions of dynamic objects, which are of key interest in applications. This leads to better optical flow with visibly more detailed occlusions and object boundaries as a result. Our unsupervised pipeline achieves 7.36\% optical flow error on the KITTI-2015 benchmark and outperforms the previous state-of-the-art 9.38\% by a wide margin. It also achieves slightly better or comparable stereo depth results. Code will be made available upon acceptance of this paper.
\end{abstract}



\doparttoc 
\faketableofcontents 

\part{} 

\section{Introduction} \label{sec:intro}

\begin{figure*}[tb]
    \centering
    \includegraphics[width=\linewidth]{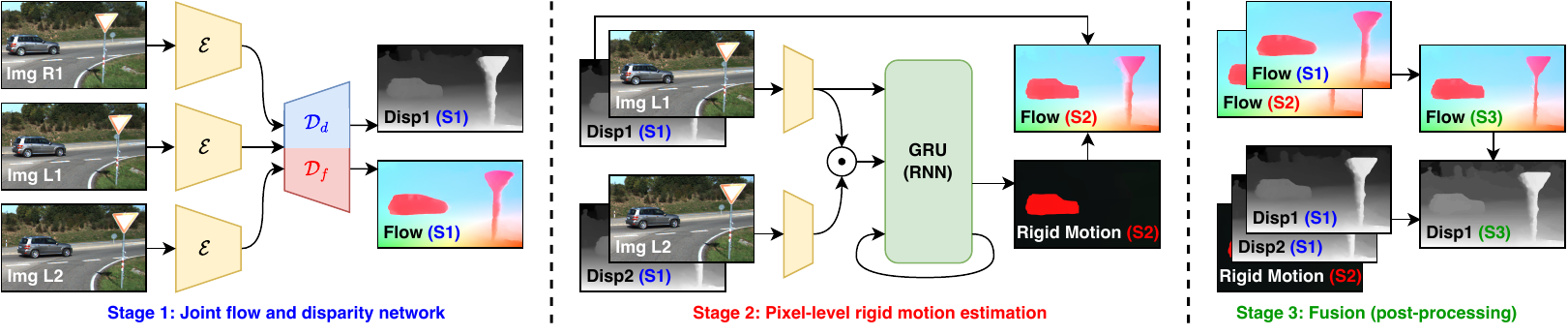}
    \caption{Pipeline overview. In Stage 1 (S1), we train a joint unsupervised network adapted from ARFlow~\cite{liu2020learning} to estimate optical flow and disparities. In Stage 2 (S2), we combine RGBD input using disparity and train RAFT3D~\cite{teed2021raft} supervised by the optical flow pseudo-label estimated in Stage 1. In Stage 3 (S3), we fuse the results from S1 and S2. Stage 3 is a pure post-processing stage with no training.}
    \label{fig:intro}
\end{figure*}

The estimation of optical flow and stereo depth are long-lasting problems in computer vision. They help intelligent systems understand 3D structure and motion in applications such as autonomous driving~\cite{geiger2012we}, virtual/augmented reality~\cite{bastug2017toward}, and robotics~\cite{dong2022towards}.

Since deep neural networks have revolutionized many traditional computer vision tasks~\cite{krizhevsky2012imagenet,girshick2015fast,he2016identity,kim2022cross,kong2023mitigating}, various supervised networks have been proposed to learn optical flow~\cite{dosovitskiy2015flownet,sun2018pwc,hur2019iterative,teed2020raft,zhang2021separable,jiang2021learning,huang2022flowformer,shi2023flowformer++,jung2023anyflow} and stereo depth~\cite{kendall2017end,guo2019group,chang2018pyramid,zhang2019ga,cheng2019learning,lipson2021raft} end-to-end. However, these systems, and especially the most recent ones~\cite{dosovitskiy2020image,carion2020end,ho2020denoising}, demand high-quality ground truth. Since annotating real data with optical flow or depth can be very expensive~\cite{yuan2022optical}, much recent interest has turned to unsupervised training~\cite{yu2016back,ren2017unsupervised}.

Inspired by traditional methods~\cite{lucas1981iterative,horn1981determining,revaud2015epicflow}, recent unsupervised flow and stereo matching networks rely on the constant brightness and smoothness assumptions to design loss functions~\cite{meister2018unflow,liu2019selflow,liu2020learning,luo2021upflow,stone2021smurf,liu2020flow2stereo}. These methods share some of the issues of traditional methods with degraded estimates due to occlusions~\cite{wang2018occlusion}, motion boundaries~\cite{yu2022unsupervised}, non-Lambertian surfaces~\cite{chen2020deep}, lack of texture~\cite{shi1994good}, and illumination changes~\cite{mohamed2014illumination}. 

To better learn optical flow and stereo depth, a natural idea from multi-task learning~\cite{ruder2017overview} is to combine information from both tasks so they can benefit from each other. For example, given a stereo video sequence, stereo disparity can be estimated from left frame $I_{{L1}}$ synchronized to right frame $I_{{R1}}$, and optical flow can be found from $I_{{L1}}$ and its temporal successor $I_{{L2}}$. Scene depth can be computed from stereo disparity given camera calibration. Flow and disparity should be consistent in terms of structural layouts and motion patterns. They both result from image matching and can therefore be estimated jointly by a unified network that reuses features and parameters across tasks~\cite{liu2020flow2stereo}.

Stereo depth can also be used to reconstruct optical flow~\cite{yin2018geonet,ranjan2019competitive}. Specifically, if objects do not deform, a number of 6-degree-of-freedom rigid motions occur in the field of view. Given camera calibration, the positions and 3D motions of all scene points can be computed, and the resulting 3D trajectories can be projected to the image plane to obtain optical flow. This reconstructed flow is available even if the point is occluded or out-of-sight in the next frame, so it can potentially complement the flow computed by 2D photometric matching.

Motivated by these close relationships between optical flow and stereo depth, many methods have explored training flow and depth together with rigid motions~\cite{yin2018geonet,zou2018df,ranjan2019competitive,luo2019every,wang2019unos,chi2021feature,jiao2021effiscene,wang2020unsupervised,liu2019unsupervised}. However, most methods assume stationary scenes and only estimate the camera motion (also known as ``egomotion'') either by traditional methods such as SfM~\cite{ullman1979interpretation,schonberger2016structure}, PnP~\cite{fischler1981random,lepetit2009ep}, and ICP~\cite{besl1992method}, or by neural networks~\cite{vijayanarasimhan2017sfm,zhou2017unsupervised}. This may cause problems for dynamic objects such as moving vehicles, which are usually of key interest in autonomous driving.
 
Some methods compute a binary segmentation mask to indicate which pixels may be problematic during flow reconstruction~\cite{ranjan2019competitive,jiao2021effiscene,wang2019unos,chi2021feature}. However, these masks are hard to learn \emph{per se}~\cite{yang2021learning}. Even when they are correct, the rigid motions of dynamic objects are typically discarded rather than corrected. We argue that a more detailed rigid motion representation can handle dynamic objects better.

To this end, we propose UFD-PRiME, an unsupervised joint model for optical flow and stereo depth inference with a specific focus on rigid motion estimation for every pixel. As shown in \cref{fig:intro}, our system has three stages (\cref{subsec:three_stages}). We first train a light-weight joint network adapted from ARFlow~\cite{liu2020learning} to obtain good initial estimates of flow and disparity (\cref{subsec:stage1}). Subsequently, we adapt RAFT-3D~\cite{teed2021raft} to generate a dense rigid motion map for flow reconstruction (\cref{subsec:stage2}). Finally, we fuse and refine our results from previous stages (\cref{subsec:stage3}). To the best of our knowledge, we are the first to introduce dense rigid motion maps to the unsupervised training of flow and stereo depth.

Our system outperforms all previous methods both quantitatively (\cref{subsec:benchmark}) and qualitatively (\cref{subsec:qual_example}). Our final stage achieves 7.36\% optical flow error on the KITTI-2015 benchmark~\cite{kitti15}, which is significantly better than previous state-of-the-art systems UPFlow~\cite{luo2021upflow} (9.38\%) and FLC~\cite{chi2021feature} (9.70\%), while maintaining marginally better stereo depths at the same time. Surprisingly, even our simple first-stage network achieves 9.01\% optical flow error, already outshining the state-of-the-art methods. Moreover, scene flow evaluations suggest that our system indeed captures accurate 3D motion information (\cref{subsec:scene_flow}). Extensive ablation studies justify the effectiveness of our current network settings (\cref{subsec:ablation}). Despite having three stages, our system runs efficiently thanks to the small network sizes (\cref{subsec:time}).

In summary, our contributions are as follows.
\begin{itemize}
    \item We propose a simple yet effective unsupervised joint network for optical flow and stereo depth that achieves state-of-the-art performance.
    \item We show the effectiveness of pixel-level rigid motion in improving optical flow results, especially on dynamic objects and occlusion regions. To the best of our knowledge, we are the first to adopt dense rigid motion maps in the unsupervised training of flow and stereo depth.
    \item We provide complete training and testing code together with our trained models at all stages (see supplementary material) for the sake of reproducibility.
\end{itemize}

\section{Related Work} \label{sec:rel_works}

\begin{figure*}[tb]
  \centering
  \begin{subfigure}{0.33\linewidth}
    \includegraphics[width=\linewidth]{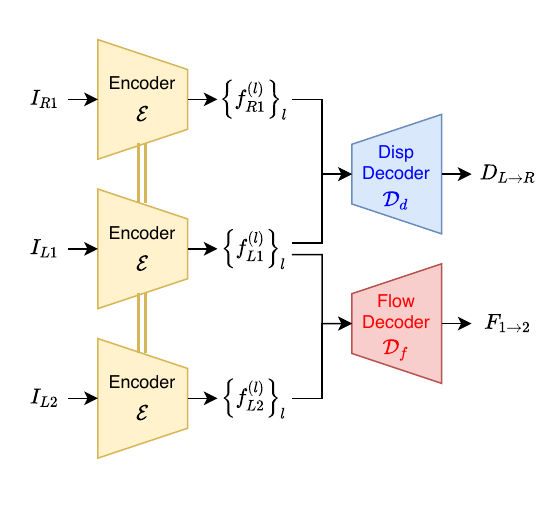}
    \caption{Network structure overview.}
    \label{fig:stage1_overview}
  \end{subfigure}
  \hfill
  \begin{subfigure}{0.66\linewidth}
    \includegraphics[width=\linewidth]{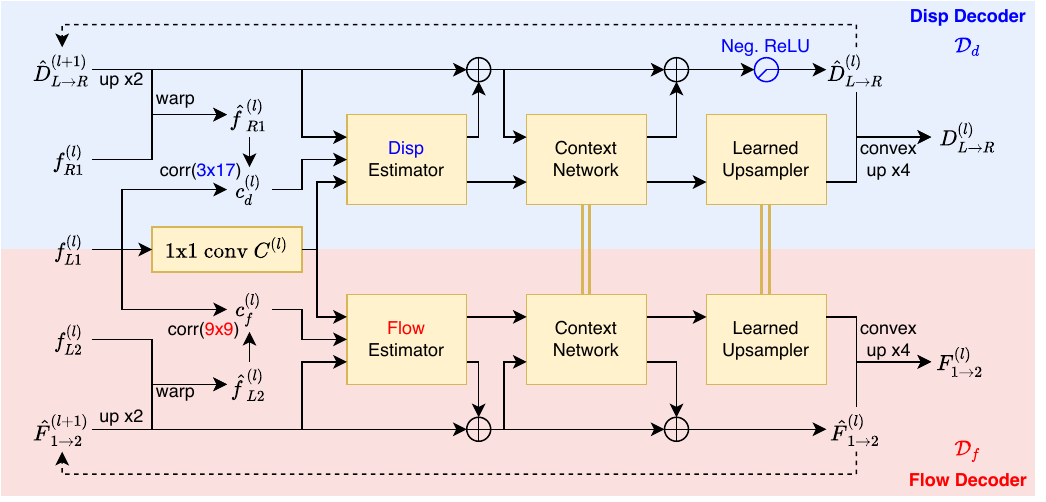}
    \caption{One iteration of the iterative decoder at the $l$th-level ($l\in\{6, 5, 4, 3, 2\}$).}
    \label{fig:stage1_dec}
  \end{subfigure}
\caption{Our Stage-1 network structure adapted from ARFlow~\cite{liu2020learning}. Double lines stand for weight sharing. See text in \cref{subsec:stage1} for explanations. See Appendix A.1. for more detailed structures.}
\label{fig:stage1}
\end{figure*}

\paragraph{Optical Flow Estimation}
Although many successful supervised optical flow methods have been proposed in recent years~\cite{sun2018pwc,hur2019iterative,teed2020raft,zhang2021separable,jiang2021learning,huang2022flowformer,shi2023flowformer++}, the unsupervised estimation of optical flow remains a challenging problem. Early unsupervised methods adopt photometric and smoothness losses as surrogates of ground truth~\cite{yu2016back,ren2017unsupervised,meister2018unflow}. Follow-up techniques have been proposed to better train flow, including occlusion masking~\cite{wang2018occlusion,meister2018unflow}, iterative refinement~\cite{hur2019iterative,liu2020learning}, learned upsampling~\cite{yuan2023semarflow,luo2021upflow}, and multi-frame fusion~\cite{janai2018unsupervised,ren2019fusion,stone2021smurf}. Self-supervised training has also shown to be effective in boosting model performance through teacher-student models~\cite{liu2019ddflow,liu2019selflow}, augmentation loss~\cite{liu2020learning}, and synthetic dataset learning~\cite{huang2023self,stone2021smurf}. We adopt ARFlow~\cite{liu2020learning} as our backbone flow network due to its simplicity and good performance.

\paragraph{Stereo Depth Estimation}
Stereo depth estimation computes the disparity between rectified stereo images. Early traditional methods develop hand-crafted matching costs~\cite{hannah1974computer,zabih1994non} and matching algorithms~\cite{kolmogorov2005convergent,barnes2009patchmatch,hirschmuller2007stereo}. Supervised CNNs have also been introduced to learn disparity using 3D cost volumes~\cite{kendall2017end,guo2019group,chang2018pyramid,zhang2019ga,cheng2019learning} and recurrent field transforms~\cite{lipson2021raft}. Due to the scarcity of ground-truth labels, many unsupervised networks have also been proposed, which aim at learning deep matching features from confident matches~\cite{zhou2017unsupervised_st, joung2019unsupervised} and disparity map smoothness~\cite{li2021unsupervised}. Depth has also been predicted from monocular images in an unsupervised manner~\cite{godard2017unsupervised,vijayanarasimhan2017sfm}. We rely on stereo depth because it is generally more stable and can generalize better.

\paragraph{3D Rigid Motion Estimation}
3D rigid motion is a compact motion representation with six degrees of freedom~\cite{forsyth2002computer} and has been extensively studied in traditional computer vision~\cite{torr1998geometric,torr1999problem,vidal2003optimal,vidal2006two,xu20193d,sawhney19943d}. Traditional methods estimate dense 3D rigid motions from frame correspondences and geometric constraints, which are sensitive to noise, especially for degenerate cases with co-plane/co-linear motions~\cite{yuan2007detecting} or degenerate camera motions~\cite{torr1999problem}.
Many neural network methods have also been proposed to learn 3D rigid motion~\cite{vijayanarasimhan2017sfm,zhou2017unsupervised,teed2021raft}. RAFT-3D~\cite{teed2021raft} mimics an optimization process with Special Euclidean Lee algebra~\cite{thurston1997three,teed2021tangent} to recurrently refine the rigid motion map, supervised by scene flow labels. We adapt from RAFT-3D due to its top supervised performance.

\paragraph{Joint Learning of Flow and Depth}




Early methods have combined flow and disparity networks using spatial-temporal consistency~\cite{lai2019bridging,liu2020flow2stereo,xu2023unifying,hur2020self,bayramli2023raft}. Other work estimates camera motion from the input frames, flow, or features to reconstruct 3D flow, which is then constrained to be consistent with 2D optical flow estimates~\cite{yin2018geonet,zou2018df,wang2020unsupervised,hur2020self}. Binary segmentation masks that separate camera motion from other dynamic motions have been either computed~\cite{luo2019every,wang2019unos,liu2019unsupervised,chi2021feature} or learned~\cite{ranjan2019competitive,jiao2021effiscene}. The results of these methods underscore the benefits or training flow and stereo depth jointly.

\section{Method} \label{sec:method}

\subsection{Problem Definition} \label{subsec:prob_def}
The input of our system is a set of two consecutive stereo RGB frame pairs $I_{{L1}}, I_{{R1}}, I_{{L2}}, I_{{R2}}\in\R^{H\times W\times 3}$, where the subscript ``L/R'' refers to left/right view and ``1/2'' refers to the first/second in temporal order. Without loss of generality, our goal is to estimate the left-view optical flow $F_{{L1}\to{L2}}\in\R^{H\times W\times 2}$ and the first-time disparity $D_{{L1}\to{R1}}\in\R^{H\times W\times 1}$. We assume the camera parameters are known, including camera intrinsics $K$ and baseline distance $b$.

\subsection{Overview of Three Stages}\label{subsec:three_stages}

As shown in \cref{fig:intro}, our system contains three stages. We first train an unsupervised network to estimate optical flow and disparity jointly (\cref{subsec:stage1}). Images and Stage-1 disparities yield RGBD inputs for a network that estimates pixel-level 3D rigid motion and is trained with Stage-1 flow as pseudo-labels. This rigid motion map is used to reconstruct dense 3D motion, which is then projected onto the image plane to yield Stage-2 optical flow (\cref{subsec:stage2}). Stage 3 fuses results from previous stages into the final outputs (\cref{subsec:stage3}).

\subsection{Stage 1: Joint Flow and Disparity Network}\label{subsec:stage1}

\paragraph{Overview}
The overall network structure is shown in \cref{fig:stage1_overview}. We first apply the same encoder $\mathcal{E}$ to extract multi-scale feature pyramids $f_{{L1}}^{(l)}, f_{{L2}}^{(l)}, f_{{R1}}^{(l)}$ on the $l$-th level ($2\leq l\leq 6$). Then, we use two different decoders $\mathcal{D}_f, \mathcal{D}_d$ to compute optical flow $F_{1\to2}$ and disparity $D_{{L}\to{R}}$.

Some previous methods such as FLC~\cite{chi2021feature} use one unified decoder to solve both tasks together. In contrast, we insist to use two separate decoders for flexibility. For instance, we can compute the backward flow and disparity $F_{2\to1}$, $D_{{R}\to{L}}$ in the same pass of the network by simply swapping the feature inputs to each decoder. This property is especially important for unsupervised training because the backward flow and disparity are required when estimating occlusion masks (based on forward-backward consistency~\cite{meister2018unflow}) used in the unsupervised photometric loss~\cite{wang2018occlusion}.

\paragraph{Shared Decoders}

Although the flow and disparity decoders can run separately, they can share weights in some modules for joint learning. Optical flow and disparity estimation are essentially both correspondence matching problems, so the decoders should work in similar ways. However, disparity estimation is a 1D search problem rather than a 2D search like optical flow. Thus, we also make changes to the disparity decoder to utilize that special property.

We adopt the flow decoder from ARFlow~\cite{liu2020learning} due to its simplicity and light weight. As shown in the lower part of \cref{fig:stage1_dec}, our flow decoder contains a warping and correlation module, a flow estimator network, a context network, as well as a learned upsampler network suggested by SemARFlow~\cite{yuan2023semarflow}. The decoder is applied recurrently to refine flow estimate $\hat F_{1\to2}^{(l)}$ starting from zero flow $\hat F_{1\to2}^{(7)}=\bm 0$ until 1/4 resolution $\hat F_{1\to2}^{(2)}$. The upsampled $F_{1\to2}^{(2)}$ is then used as the final output of the flow decoder. We refer readers to the original papers~\cite{liu2020learning, yuan2023semarflow} for more details.

The disparity decoder structure is shown in the upper part of \cref{fig:stage1_dec}. We copy the same structure from the flow decoder except for the following small changes.
\begin{itemize}
    \item We change the window size of the pairwise correlation module from $9\times 9$ to $3\times 17$ to make the disparity decoder focus more along the horizontal direction.
    \item We use a new disparity estimator to replace the flow estimator network since the number of input channels has changed as we change the correlation module.
    \item We apply negative ReLU, which filters only negative values, before upsampling to make sure our left-to-right disparity output is negative.
    \item We add a redundant $y$-channel of all zeros to the estimated 1D disparity $\hat D_{L\to R}^{(l)}$ to make it 2D so that we can reuse the same context network and learned upsampler.
\end{itemize}

\paragraph{Loss} We apply the same photometric and augmentation loss as in ~\cite{yuan2023semarflow} for both flow and disparity, which are then linearly combined with weights $w_f=0.7, w_d=0.3$.

\subsection{Stage 2: Pixel-Level Rigid Motion Estimation}\label{subsec:stage2}

Many previous methods have shown successful examples of reconstructing optical flow using depth maps and the 6-DoF rigid motion~\cite{jiao2021effiscene, luo2019every, wang2019unos, liu2019unsupervised, chi2021feature}. Nevertheless, most methods only estimate the rigid motion of the camera, assuming that the whole scene is stationary. This is problematic for dynamic objects such as moving vehicles.

\paragraph{Pixel-Level Rigid Motion Map}
In contrast, we explore estimating rigid motion for every rigid body in the view to enhance flow reconstruction. Since the number of objects in each frame is variable, inspired by RAFT-3D~\cite{teed2021raft}, we represent rigid motion at the pixel-level as a dense map $T\in {SE(3)}^{H\times W}$, where $SE(3)$ is the 3D Special Euclidean group for rigid motions~\cite{hartley2003multiple}. The rigid motion of each pixel $T(\bm x)$ is defined as the rigid motion of the object to which that pixel belongs. This map not only contains full rigid motion information of the scene, but also implies a segmentation of rigid bodies.

\paragraph{Reconstructing Flow from Disparity and Rigid Motion} Suppose $(x, y)$ is a point from the reference frame $I_{{L1}}$, and its disparity has been estimated as $D$. We can compute its depth $Z\in\R^+$ and its 3D coordinates $\bm X\in\R^3$ by
\begin{equation}\label{eq:depth}
Z = \frac{f_xb}{|D|}, \quad \bm X = ZK^{-1}\left[
\begin{array}{c}
x \\
y \\
1
\end{array}
\right],
\end{equation}
where $K\in\R^{3\times 3}$ is the camera intrinsics matrix, $f_x$ is the horizontal focal distance of the camera, and $b$ is the baseline distance between left and right cameras~\cite{forsyth2002computer}.

Suppose the point belongs to a rigid body that has rigid motion $(R, \bm t)$. The coordinates of that point in the next camera coordinates system will be $\bm X' = R\bm X+\bm t$, which can be projected back to the frame by
\begin{equation}\label{eq:motion}
Z'\left[
\begin{array}{c}
x' \\
y' \\
1
\end{array}
\right] = K(R\bm X+\bm t),
\end{equation}
where $(x', y')$ is the corresponding position and $Z'$ is its new depth in the new frame. The reconstructed optical flow is then $(x'-x, y'-y)$. Note that we only update the optical flow in Stage 2 with no refinement on disparities.

\paragraph{Network} We borrow RAFT-3D~\cite{teed2021raft} as our network structure. RAFT-3D is a supervised scene flow network that takes RGBD inputs and estimates rigid motion maps using a Dense-SE3 layer~\cite{teed2021raft}. The rigid motion maps are recurrently refined and used to reconstruct scene flow, which is then supervised by labels in an end-to-end manner.

Different from RAFT-3D, our system only takes RGB inputs without depths, as defined in \cref{subsec:prob_def}. We also do not have the scene flow labels for supervision. Therefore, we use our Stage-1 model to generate depth inputs and flow pseudo-labels to train the model. 

Since our Stage-1 outputs may not be reliable in occlusion regions, we also estimate their occlusion masks through forward-backward consistency check~\cite{meister2018unflow}. The occlusion regions for both disparities and flows are masked out in the loss, so we only penalize at places where both our predictions and pseudo-labels are reliable. More details are included in Appendix A.2.

\paragraph{Loss}
RAFT-3D~\cite{teed2021raft} is originally trained with a supervised scene flow loss, which can be computed using our pseudo-labels from Stage-1 model inference. In addition, we add a smoothness loss to better constrain our rigid motion map. Specifically, we compute the first-order gradient of the 6D rigid motion map and take its L1 norm as our smoothness loss, which is applied to estimates at all decoder iterations in accordance with the original RAFT-3D loss. We expect that the rigid motion for occlusion regions can be imputed based on local smoothness, which allows us to reconstruct occluded flow accurately.

\begin{table*}[tb]
\centering
\begin{tabular}{l|cc|cc|ccccc}
\thickhline
\multicolumn{1}{c|}{\multirow{3}{*}{Methods}} &  \multirow{3}{*}{Joint?}         & \multirow{3}{*}{Stereo?}     & \multicolumn{2}{c|}{KITTI-2012}        & \multicolumn{5}{c}{KITTI-2015}                                     \\ \cline{4-10} 
\multicolumn{1}{c|}{}                        &  &  & \multicolumn{1}{c|}{train}   & test    & \multicolumn{4}{c|}{train}                                & test   \\
\multicolumn{1}{c|}{}                        &               &         & \multicolumn{1}{c|}{EPE} & EPE & EPE-all & EPE-noc & EPE-occ & \multicolumn{1}{c|}{Fl-all} & Fl-all \\ \hline
SelFlow~\cite{liu2019selflow}                &                &         & \multicolumn{1}{c|}{1.69}    & 2.2     & 4.84    &  -       &   -      & \multicolumn{1}{c|}{-}       & 14.19  \\
ARFlow~\cite{liu2020learning}                &                &         & \multicolumn{1}{c|}{1.44}    & 1.8     & 2.85    &  -       &  -       & \multicolumn{1}{c|}{-}       & 11.80  \\
UPFlow~\cite{luo2021upflow}                  &                &         & \multicolumn{1}{c|}{1.27}    & {1.4}     & 2.45    &  -      &    -     & \multicolumn{1}{c|}{-}       & 9.38   \\ \hline
DF-Net~\cite{zou2018df}                      & \cmark              &         & \multicolumn{1}{c|}{3.54}    & 4.4     & 8.98    &   -      &     -    & \multicolumn{1}{c|}{26.01}  & 25.70  \\
CC-uft~\cite{ranjan2019competitive}          & \cmark              &         & \multicolumn{1}{c|}{-}        &    -     & 5.66    & -        &   -      & \multicolumn{1}{c|}{20.93}  & 25.27  \\
EPC++ (mono)~\cite{luo2019every}             & \cmark              &         & \multicolumn{1}{c|}{2.30}    & 2.6     & 5.84    &   -      &    -     & \multicolumn{1}{c|}{-}       & 21.56  \\ \hline
EPC++ (stereo)~\cite{luo2019every}           & \cmark              & \cmark       & \multicolumn{1}{c|}{1.91}    & 2.2     & 5.43    &  -       &    -     & \multicolumn{1}{c|}{-}       & 20.52  \\
UnOS~\cite{wang2019unos}                     & \cmark              & \cmark       & \multicolumn{1}{c|}{1.92}    & -     & 5.58    &  -       &   -      & \multicolumn{1}{c|}{-}       & 18.00  \\
UnRigidFlow~\cite{liu2019unsupervised}                     & \cmark              & \cmark       & \multicolumn{1}{c|}{1.64}    & 1.8     & 5.19    &  -       &   -      & \multicolumn{1}{c|}{-}       & 11.66  \\
Flow2Stereo~\cite{liu2020flow2stereo}        & \cmark              & \cmark       & \multicolumn{1}{c|}{1.45}    & 1.7     & 3.54    & 2.12    &     -    & \multicolumn{1}{c|}{-}       & 11.10  \\
EffiScene~\cite{jiao2021effiscene}           & \cmark              & \cmark       & \multicolumn{1}{c|}{1.68}    &   -      & 4.20    &  -       &   -      & \multicolumn{1}{c|}{14.31}  & 13.08  \\
FLC~\cite{chi2021feature}                    & \cmark              & \cmark       & \multicolumn{1}{c|}{{1.25}}    & 1.5     & 2.35    & {1.57}    & 6.68    & \multicolumn{1}{c|}{9.09}   & 9.70   \\ \hline
\textbf{Ours (Stage 1)}                      & \cmark              & \cmark       & \multicolumn{1}{c|}{1.26}    & 1.5     & {2.31}    & 1.70    & {5.31}    & \multicolumn{1}{c|}{{7.93}}   & {9.01}   \\
\textbf{Ours (Stage 2)}                      & \cmark              & \cmark       & \multicolumn{1}{c|}{{1.04}}    & \textbf{1.2}     & {2.04}    & {1.49}    & \textbf{4.78}    & \multicolumn{1}{c|}{{6.87}}   & {7.63}   \\ 
\textbf{Ours (Stage 3)}                      & \cmark              & \cmark       & \multicolumn{1}{c|}{\textbf{1.02}}    & \textbf{1.2}     & \textbf{1.99}    & \textbf{1.42}    & {4.87}    & \multicolumn{1}{c|}{\textbf{6.65}}   & \textbf{7.36}   \\ 
\thickhline
\end{tabular}
\caption{Optical flow test errors on KITTI benchmarks (EPE/px and Fl/\%). Metrics evaluated at ``all'' (all pixels), ``noc'' (non-occlusions), and ``occ'' (occlusions). The ``Joint?'' column tells whether the method conducts joint training of flow and depth. The ``Stereo?'' column tells whether the method uses stereo camera inputs. ``-'' refers to unavailable data. The best result in each column is printed in bold.}
\label{tab:flow_res}
\end{table*}

\begin{table*}[tb]
\centering
\begin{tabular}{l|ccccccc}
\thickhline
\multicolumn{1}{c|}{\multirow{2}{*}{Methods}} & \multicolumn{4}{c|}{Error (Lower is Better)}                                                                                           & \multicolumn{3}{c}{Accuracy (Higher is Better)}    \\ 
\multicolumn{1}{c|}{}                         & Abs Rel        & Sq Rel         & RMSE           & \multicolumn{1}{c|}{RMSE-log}              & $\delta<1.25$  & $\delta<1.25^2$ & $\delta<1.25^3$ \\ \hline
Godard et al. ~\cite{godard2017unsupervised}                      & 0.068          & 0.835          & 4.392          & \multicolumn{1}{c|}{0.146}                       & 0.942          & 0.978           & {0.989}  \\
UnOS~\cite{wang2019unos}                      & 0.049          & 0.515          & 3.404          & \multicolumn{1}{c|}{0.121}                       & 0.965          & 0.984           & \textbf{0.992}  \\
UnRigidFlow~\cite{liu2019unsupervised}        & 0.051          & 0.532          & 3.780          & \multicolumn{1}{c|}{0.126}                       & 0.957          & 0.982           & 0.991           \\
EffiScene~\cite{jiao2021effiscene}            & 0.049          & 0.522          & 3.461          & \multicolumn{1}{c|}{0.120}                       & 0.961          & 0.984           & \textbf{0.992}  \\
FLC~\cite{chi2021feature}                     & \textbf{0.047} & \textbf{0.394} & \textbf{3.358} & \multicolumn{1}{c|}{0.119}           & -              & -               & -               \\ \hline
\textbf{Ours (Stage 1 \& Stage 2)}             & 0.048          & 0.574          & 3.616          & \multicolumn{1}{c|}{\textbf{0.118}}        & {0.970} & \textbf{0.986}  & \textbf{0.992}  \\ 
\textbf{Ours (Stage 3)}             & \textbf{0.047}          & 0.565          & 3.588          & \multicolumn{1}{c|}{\textbf{0.118}}        & \textbf{0.971} & \textbf{0.986}  & \textbf{0.992}  \\ 
\thickhline
\end{tabular}
\caption{Stereo depth evaluation on KITTI-2015 train set compared with other unsupervised joint training of flow and stereo depth methods. Note that we do not refine depth on Stage 2, so Stage 1 and 2 share the same results. The best result in each column is printed in bold.}
\label{tab:depth_res}
\end{table*}

\subsection{Stage 3: Fusion and Post-Processing}\label{subsec:stage3}

We now have two different optical flow estimates based on photometric (Stage 1) and geometric (Stage 2) constraints, so we find a way to fuse them. Also, our disparity is not refined in Stage 2, so we refine it here based on the dense rigid motion maps.

\paragraph{Flow Fusion} The previous stages estimate flow in two different ways, namely 2D photometric matching (Stage 1) and 3D motion reconstruction (Stage 2). These flows are reliable in different regions, so we fuse them in light of that.

We reuse the occlusion masks computed in Stage 2 to define reliable pixels for both flows. For Stage-1 flow , we define flow on non-occluded pixels as reliable. For Stage-2 flow, since it is reconstructed from the disparity map, we adopt the disparity occlusion mask and define its non-occluded pixels as reliable. Note that the flow occlusion regions are usually larger than disparity occlusions, unless the motion is very small. This indicates that our Stage-2 flow is generally more reliable than Stage 1 in typical scenarios.

To fuse the two flows, we examine their reliabilities at each pixel. If both flows are reliable, we take the average. If exactly one of them is reliable, we copy that reliable estimate. If none of them is reliable, we stick to Stage-2 flow due to its better overall performance.

\paragraph{Disparity Refinement} We aim to refine disparities given the rigid motion map and flow. For a pixel $(x_1, y_1)\in I_{{L1}}$, its correspondence $(x_2, y_2)\in I_{{L2}}$ can be found using the fused flow above. Similar as shown in ~\cref{eq:motion,eq:depth}, we can un-project the pixels to 3D camera coordinates and find their relationships from rigid motion $(R, \bm t)$ as follows,
\begin{equation}\label{eq:disp_refine}
\frac{f_xb}{|D_2|}K^{-1}\left[
\begin{array}{c}
x_2 \\
y_2 \\
1
\end{array}
\right]
= 
R\left(\frac{f_xb}{|D_1|}K^{-1}\left[
\begin{array}{c}
x_1 \\
y_1 \\
1
\end{array}
\right]\right) + \bm t,
\end{equation}
where $D_1, D_2$ are the disparities of the same point in $I_{{L1}}, I_{{L2}}$. Our current estimates $\hat D_1, \hat D_2$ can be retrieved from Stage-1 model outputs (flow warping needed to compute $\hat D_2$), and we optimize $\delta_1=D_1-\hat D_1, \delta_2=D_2-\hat D_2$ so that \cref{eq:disp_refine} holds. We can rewrite  \cref{eq:disp_refine} as
\begin{equation}\label{eq:depth_linear_eq}
\frac{1}{|\hat D_1+\delta_1|}\bm\alpha_1 + \frac{1}{|\hat D_2+\delta_2|}\bm\alpha_2 = \bm t,
\end{equation}
where we denote constant vectors $\bm\alpha_1, \bm\alpha_2\in\R^3$ as
\begin{equation}
\bm\alpha_1 = -f_xbRK^{-1}
\left[
\begin{array}{c}
x_1 \\
y_1 \\
1
\end{array}
\right], \bm\alpha_2 = f_xbK^{-1}
\left[
\begin{array}{c}
x_2 \\
y_2 \\
1
\end{array}
\right].
\end{equation}
To linearize \cref{eq:depth_linear_eq}, we do first-order Taylor's expansions for $\delta_1, \delta_2$ around zero and since $\hat D_1, \hat D_2\leq 0$, we get
\begin{equation}
\left(-\frac{1}{\hat D_1} + \frac{1}{\hat D_1^2}\delta_1\right)\bm\alpha_1 + \left(-\frac{1}{\hat D_2} + \frac{1}{\hat D_2^2}\delta_2\right)\bm\alpha_2 = \bm t,
\end{equation}
which is an over-determined linear system in $\delta_1, \delta_2$ and can be solved in the least-squares sense in closed form~\cite{anton2013elementary}.
\section{Experiments} \label{sec:expr}

\begin{figure*}[tb]
    \centering
    \includegraphics[width=\linewidth]{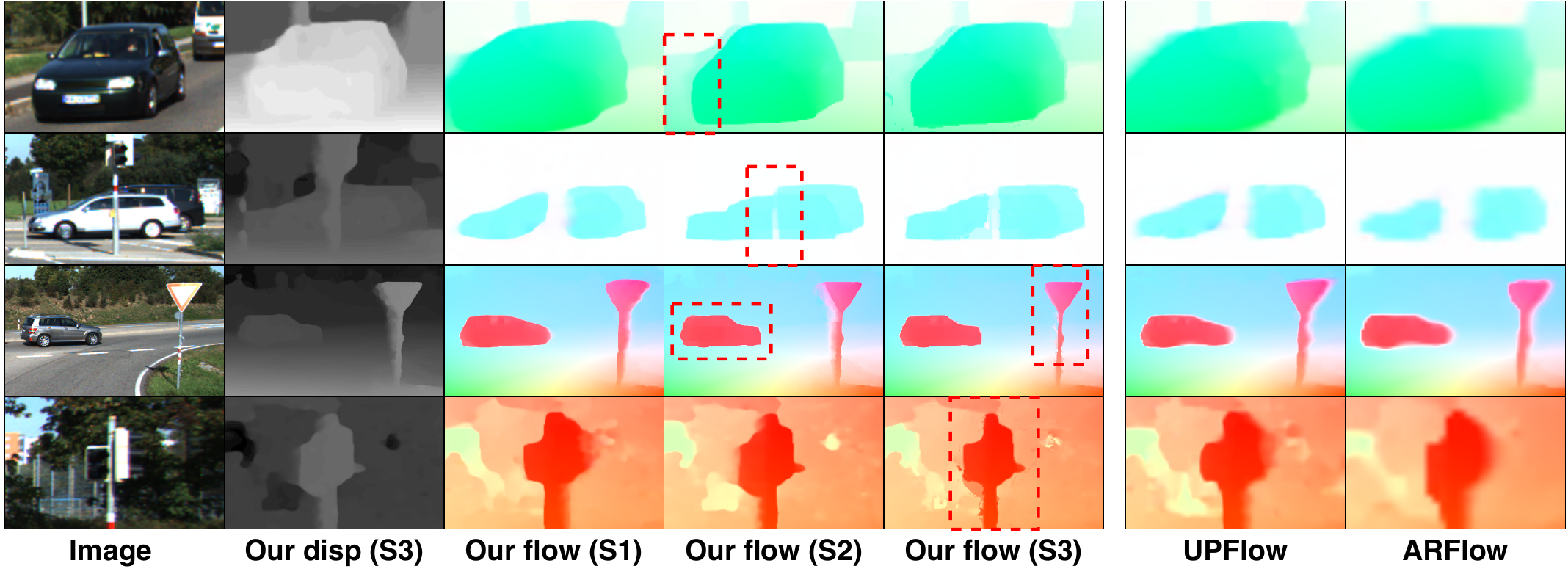}
    \caption{Qualitative results on KITTI test set (cropped from sample \#103, 94, 78, 10) compared with UPFlow~\cite{luo2021upflow} and ARFlow~\cite{liu2020learning}. Each row shows one sample. We only show the left-view first input frame. ``S1'', ``S2'', and ``S3'' refer to Stage 1, 2, and 3. See the text in \cref{subsec:qual_example} for explanations. For more qualitative examples, see Appendix B.2.}
    \label{fig:qual_examples}
\end{figure*}


\subsection{Datasets}
We experiment on KITTI datasets~\cite{kitti12,kitti15}. For all stages, our models are first trained on KITTI raw frames (25.2k samples) and then fine-tuned on KITTI multi-view extensions (6.7k samples), as suggested by ARFlow~\cite{liu2020learning}. We validate our model using KITTI-2015~\cite{kitti15} and 2012~\cite{kitti12} training sets (around 200 samples for each). 

\subsection{Implementation Details}
We train our Stage 1 and 2 PyTorch~\cite{NEURIPS2019_9015} models (see code in the supplementary material) with the Adam optimizer~\cite{kingma2014adam} with batch size 4. Stage 3 requires no training.

In Stage 1, we adopt the refined schedule from SemARFlow~\cite{yuan2023semarflow} and train on raw frames for 100k iterations with a fixed learning rate 2e-4 and then on the multi-view extension set for another 100k iterations with the OneCycleLR scheduler~\cite{smith2019super} (max learning rate 4e-4). In Stage 2, we apply the same scheduler but adjust the iterations as in RAFT-3D~\cite{teed2021raft}. We first train on raw for 200k iterations and then on multi-view for 50k iterations. All other implementation details are the same as in the originals (ARFlow~\cite{liu2020learning} for Stage 1 and RAFT-3D~\cite{teed2021raft} for Stage 2).

We augment training data with appearance transformations (brightness, contrast, saturation, hue, gaussian blur, \etc) and randomly swap the input images both in time ($I_{L1}$ and $I_{L2}$) and space ($I_{L1}$ and $I_{R1}$). With spatial swaps we also flip the images horizontally to ensure negative disparity. The inputs are resized to $384\times 1280$ in Stage 1 but cropped to $256\times 832$ in Stage 2 to save memory.

\begin{table*}[tb]
\centering
\begin{tabular}{l|ccc|ccc|ccc|ccc}
\thickhline
\multicolumn{1}{c|}{\multirow{2}{*}{Methods}} & \multicolumn{3}{c|}{Disparity 1} & \multicolumn{3}{c|}{Disparity 2} & \multicolumn{3}{c|}{Optical Flow} & \multicolumn{3}{c}{Scene Flow} \\
\multicolumn{1}{c|}{}                              & bg        & fg        & all      & bg        & fg        & all      & bg        & fg         & all      & bg       & fg       & all      \\ \hline
UnOS~\cite{wang2019unos}                              & 5.10      & 14.55      & 6.67     & 9.61      & 24.28      & 12.05     & 16.93      & 23.34       & 18.00     & 19.70     & 35.43    & 22.32     \\ \hline
Ours (Stage 1)                                       & \textbf{3.65}      & \textbf{15.04}     & \textbf{5.55}     & 14.59     & \textbf{20.09}     & 15.50    & 7.91      & \textbf{14.52}      & 9.01     & 18.13    & 28.26    & 19.82    \\
Ours (Stage 2)                                     & \textbf{3.65}      & \textbf{15.04}     & \textbf{5.55}     & \textbf{5.30}      & 20.50     & \textbf{7.83}     & \textbf{5.96}      & 15.96      & \textbf{7.63}     & \textbf{7.64}     & \textbf{26.25}    & \textbf{10.74}    \\ \thickhline
\end{tabular}
\caption{Scene flow evaluation of our method before and after pixel-level rigid motion estimation (Stage 2) on KITTI-2015 test benchmark. All metrics are error percentages (/\%). We do not refine Disparity 1 in Stage 2, so its results are the same.}
\label{tab:sf_res}
\end{table*}

\begin{table*}[tb]
\centering
\begin{tabular}{cc|cccc|cccc}
\thickhline
\multirow{2}{*}{$w_f$} & \multirow{2}{*}{$w_d$} & \multicolumn{4}{c|}{Optical Flow Error}  & \multicolumn{4}{c}{Depth Error}                  \\
                       &                        & EPE-all & EPE-noc & EPE-occ & Fl-all & Abs Rel & Sq Rel & RMSE   & RMSE-log  \\ \hline
1                      & 0                      & 2.38   & 1.73   & 5.56   & 8.02 & -    & - & - & -      \\ \hline
0.9                    & 0.1                     & 2.38   & 1.74   & 5.52   & 8.07 & 0.049    & 0.618   & 3.692  & 0.120     \\
0.7*              & 0.3*              & \textbf{2.31}   & \textbf{1.71}   & 5.31   & \textbf{7.93} & \textbf{0.048}    & \textbf{0.574}   & \textbf{3.616}  & \textbf{0.118}      \\
0.5                    & 0.5                    & 2.34   & 1.73   & \textbf{5.30}   & 7.94 & \textbf{0.048}    & 0.593   & 3.628  & \textbf{0.118}      \\
0.3                    & 0.7                    & 2.38   & 1.77   & 5.32   & 8.10   & 0.050    & 0.695   & 3.672  & 0.119      \\
0.1                    & 0.9                    & 36.92  & 27.95  & 73.69  & 83.96 & 0.050    & 0.596   & 3.658  & 0.120     \\ \hline
0                      & 1                      & -  & -  & -  & - & 0.050    & 0.586   & 3.646  & 0.120     \\
\thickhline
\end{tabular}
\caption{Ablation study (Stage 1): KITTI-2015 validation errors on optical flow and stereo depth using different balancing weights $w_f$, $w_d$. ``-'' means not applicable. * marks our final setting. The best result in each column is printed in bold.}
\label{tab:ablate_1}
\end{table*}

\setlength{\tabcolsep}{5.5pt}
\begin{table*}[tb]
\centering
\begin{tabular}{ccc|c|cccc|cccc}
\thickhline
\multicolumn{3}{c|}{Shared?} & \multirow{2}{*}{Disp. corr.} & \multicolumn{4}{c|}{Optical Flow Error}    & \multicolumn{4}{c}{Depth Error}               \\
Flow   & Context   & Up          &                        & EPE-all & EPE-noc & EPE-occ & Fl-all    & Abs Rel & Sq Rel & RMSE   & RMSE-log  \\ \hline
\cmark     & \cmark        & \cmark      & $9\times9$                           & 2.37    & 1.71    & 5.51    & 8.06    & 0.051   & 0.754  & 3.785 & 0.122       \\
       & \cmark        & \cmark      & $9\times9$                           & 2.34    & 1.71    & 5.32    & 7.96    & 0.048   & 0.588  & 3.613 & 0.118       \\
       & \cmark*        & \cmark*      & $3\times17$*                           & \textbf{2.31}    & \textbf{1.70}    & \textbf{5.31}    & \textbf{7.93} & 0.048   & 0.574  & 3.616 & 0.118       \\
       &           & \cmark      & $3\times17$                           & 2.34    & 1.71    & 5.32    & 8.03    & 0.048   & 0.593  & 3.625 & 0.118       \\
       &           &         & $3\times17$                           & 2.36    & 1.72    & 5.39    & 8.02    & \textbf{0.047}   & \textbf{0.555}  & \textbf{3.565} & \textbf{0.117}       \\ \thickhline
\end{tabular}
\caption{Ablation study (Stage 1): KITTI-2015 train set results when sharing different modules in the decoder. ``Flow'': flow/disp estimator module; ``Context'': context network module; ``Up'': the learned upsampler module; ``Disp. corr.'': the window size used in disparity correlation. * marks our final setting. The best result in each column is printed in bold.}
\label{tab:ablate_2}
\end{table*}
\setlength{\tabcolsep}{6pt}

\setlength{\tabcolsep}{5pt}
\begin{table}[tb]
\begin{tabular}{cc|c|cccc}
\thickhline
\multicolumn{2}{c|}{Options} & Train  & \multicolumn{4}{c}{Test}        \\
Mask     & Smooth     & Fl-all & Fl-all & Fl-noc & Fl-bg & Fl-fg \\ \hline
            &                & 8.47  &  9.53      &  6.93      & 8.38      & 15.25      \\
\cmark          &                & 7.19  & 7.87   & 6.42   & 6.56  & \textbf{14.42} \\
\cmark*          & \cmark*             & \textbf{6.87}  & \textbf{7.63}   & \textbf{6.29}   & \textbf{5.96}  & {15.96} \\ 
\thickhline
\end{tabular}
\caption{Ablation study (Stage 2): KITTI-2015 train and test results using different options. * marks our final setting. ``Mask'' means applying reliability masks in training. ``Smooth'' means adding smoothness loss to the rigid motion embedding. The best result in each column is printed in bold.}
\label{tab:ablate_3}
\end{table}
\setlength{\tabcolsep}{6pt}

\begin{figure}[tb]
    \centering
    \includegraphics[width=\linewidth]{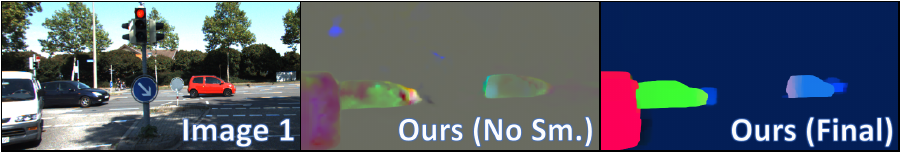}
    \caption{Visual comparisons of the estimated dense rigid motion fields with and without smoothness loss (KITTI test sample \#8).}
    \label{fig:rigid_demo}
\end{figure}

\subsection{Benchmark Test Results}\label{subsec:benchmark}

\paragraph{Optical Flow} As shown in \cref{tab:flow_res}, even our Stage 1 results outperform all previous state-of-the-art unsupervised methods including FLC\cite{chi2021feature} and UPFlow~\cite{luo2021upflow} on all metrics on KITTI-2015~\cite{kitti15} and 2012~\cite{kitti12}. Through rigid motion estimation in Stage 2, our reconstructed flow results improve even further. After leveraging and fusing flows in the first two stages, our Stage 3 finally achieves the best test errors (7.36\%) on KITTI-2015, a 20\%+ decrease compared with the state-of-the-art UPFlow~\cite{luo2021upflow} (9.38\%). These results strongly indicate that all our stages contribute significantly to improving flow results.

\paragraph{Stereo Depth} Since we do not refine disparity in Stage 2, disparities from Stage 1 and 2 are the same. \cref{tab:depth_res} shows that our Stage 1 and 2 results outperform the state of the art on most evaluation metrics (still comparable if not better). The Stage-3 refinement also improves the squared relative error and RMSE, indicating better depths on far-away objects. Stereo matching is known to be an easier problem than optical flow, so it has less margin for improvement.

\subsection{Qualitative Examples}\label{subsec:qual_example}

The examples in \cref{fig:qual_examples} illustrate how each stage works compared with UPFlow~\cite{luo2021upflow} (state-of-the-art) and ARFlow ~\cite{liu2020learning} (the backbone of our Stage-1 model).

Our Stage-1 results are visually comparable with the state-of-the-art methods. After applying pixel-level rigid motion estimation in Stage 2, our method handles occlusions around moving cars better, whether they are in the foreground (Row 1 and 3) or occluded by other objects like poles (Row 2). This is consistent with our claim that the occluded flow can be reconstructed based on stereo depth, which is generally more reliable.

Rows 3 and 4 in \cref{fig:qual_examples} show that our Stage-3 refinement can help sharpen thin foreground objects like traffic lights and poles. In this example, the traffic light is moving to the right. The Stage-1 flow is only sharp on the left side due to occlusions on the right, whereas the Stage-2 flow is only sharp on the right due to disparity occlusions on the left. By analyzing the reliable regions of each output, our fused Stage-3 output is sharp on both sides of the object. This illustrates a typical way in which Stage 3 improves, \ie by combining the flow results from photometric (Stage 1) and geometric (Stage 2) constraints that are usually reliable in different regions.

\subsection{Scene Flow Evaluations}\label{subsec:scene_flow}
As in RAFT-3D~\cite{teed2021raft}, we evaluate our Stage-2 rigid motion estimates via scene flow due to lack of dense rigid-motion ground truth. Our Stage 1 does not output ``Disparity 2'' (the changed disparity of points from the first frame), so we approximate it by warping our first-pair disparity with flow. \cref{tab:sf_res} shows that our Stage-2 rigid motion estimation very much improves scene flow results, suggesting that it captures dense 3D rigid motions well.

\subsection{Ablation Studies}\label{subsec:ablation}
\paragraph{Balancing Optical Flow and Disparity in Stage 1} \cref{tab:ablate_1} shows that $w_f=0.7, w_d=0.3$ are the best balancing weights for flow and disparity losses in Stage 1. We also compare with settings where we train on either flow or disparity alone using the same network and show that our joint network benefits both tasks.

\paragraph{Different Options of Shared Decoders in Stage 1}
We examine different options for how the two decoders in Stage 1 share model weights, from the most shared (Row 1) to the least shared setting (Row 5). \cref{tab:ablate_2} shows that our final setting has the best flow results while maintaining comparable depth evaluations.

\paragraph{Changes to RAFT-3D in Stage 2}\cref{tab:ablate_3} shows that our proposed changes in Stage 2 are necessary to optimize RAFT-3D~\cite{teed2021raft} in the unsupervised setting. 
Moreover, \cref{fig:rigid_demo} visualizes the 6 DoF rigid motion field in RGB colors after reducing dimensionality to 3 with PCA~\cite{pearson1901liii}. The added smoothness loss clearly helps separate all rigid bodies in the frame. This also implies that our dense rigid motion map can potentially be used to generate rigid body segmentation masks, which we discuss in Appendix B.3.

\subsection{Time Efficiency}\label{subsec:time} 
We time our stages and show that they run efficiently. We infer input samples of dimension $376\times 1242$ on one NVIDIA GeForce RTX 2080 Ti GPU. Our Stage-1 model contains 3.2 million parameters and can infer both flow and disparity together in $0.077(\pm0.001)$ seconds. Our Stage-2 model contains 6.0 million parameters and can perform dense rigid motion estimation and flow reconstruction in $0.471(\pm 0.052)$ seconds. Our Stage 3 is a simple post-processing stage and runs instantly on CPUs.

\section{Conclusion} \label{sec:con}

We propose UFD-PRiME, a system for the joint unsupervised training of optical flow and stereo depth using pixel-level rigid motion estimation. We first design a simple yet effective unsupervised network based on ARFlow~\cite{liu2020learning} to train flow and disparity jointly, where we use separate decoders with partial weight-sharing modules to handle the two tasks. Then, we train a pixel-level rigid motion estimation network adapted from RAFT-3D~\cite{teed2021raft} to estimate dense 3D rigid motion maps, which are then used to reconstruct flow once again given depth. Finally, we fuse and refine flow and disparities by reliability analysis and stereo geometry. Our method outperforms all previous methods significantly on optical flow errors on the KITTI benchmarks~\cite{kitti12,kitti15}, especially on occlusion regions and around dynamic objects, while maintaining marginally better stereo depth evaluations.

\paragraph{Limitations} 
Estimating rigid motion for each pixel alone can be sensitive. For each 2D motion vector observed in the image plane, there are many different 3D rigid motions that can generate the same 2D projected motion. In our system, we use a smoothness loss on the rigid motion map to alleviate this issue. A possibly better solution is to aggregate object and instance level information so that we can assign the same rigid motion to every pixel on the same rigid body. We leave this study for future work.

\clearpage
\newpage

\appendix
\addcontentsline{toc}{section}{Appendix} 
\part{Appendix} 
\parttoc 

\appendix

\tableofcontents

\section{Method Details}

\subsection{Stage 1}

\begin{figure*}[tb]
  \centering
  \begin{subfigure}{0.6\linewidth}
    \centering
    \includegraphics[width=\linewidth]{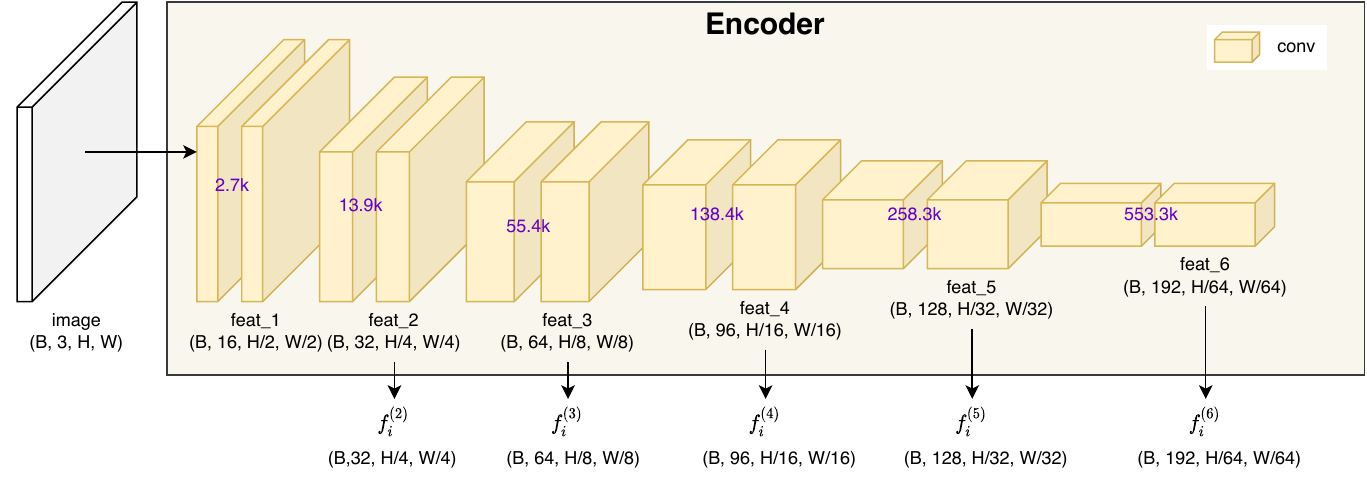}
    \caption{Encoder structure (Stage 1)}
    \label{fig:enc_app}
 \end{subfigure}
 \par\bigskip
  \begin{subfigure}{0.95\linewidth}
    \centering
    \includegraphics[width=\linewidth]{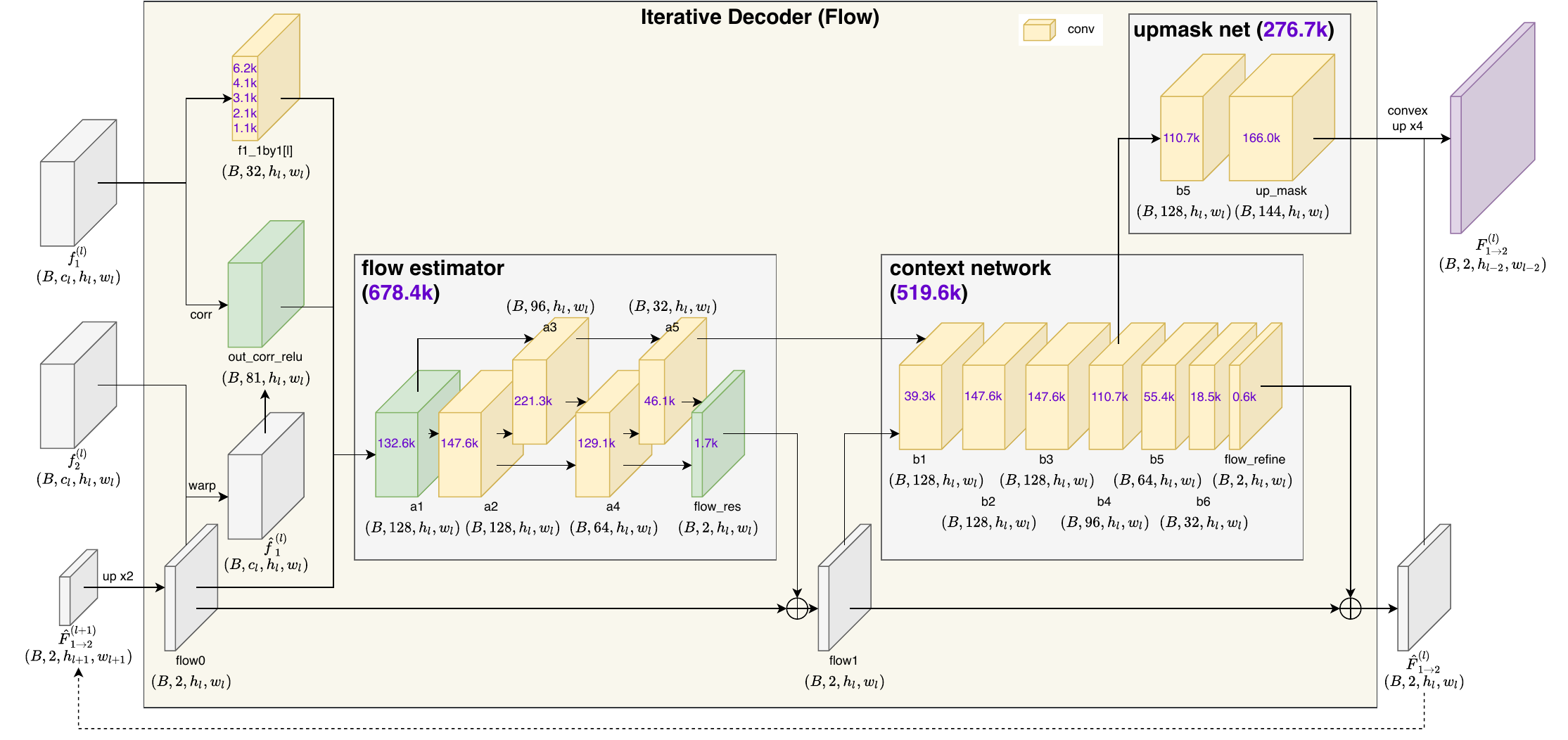}
    \caption{Flow decoder structure (Stage 1)}
    \label{fig:dec_f_app}
 \end{subfigure}
 \par\bigskip
  \begin{subfigure}{0.95\linewidth}
    \centering
    \includegraphics[width=\linewidth]{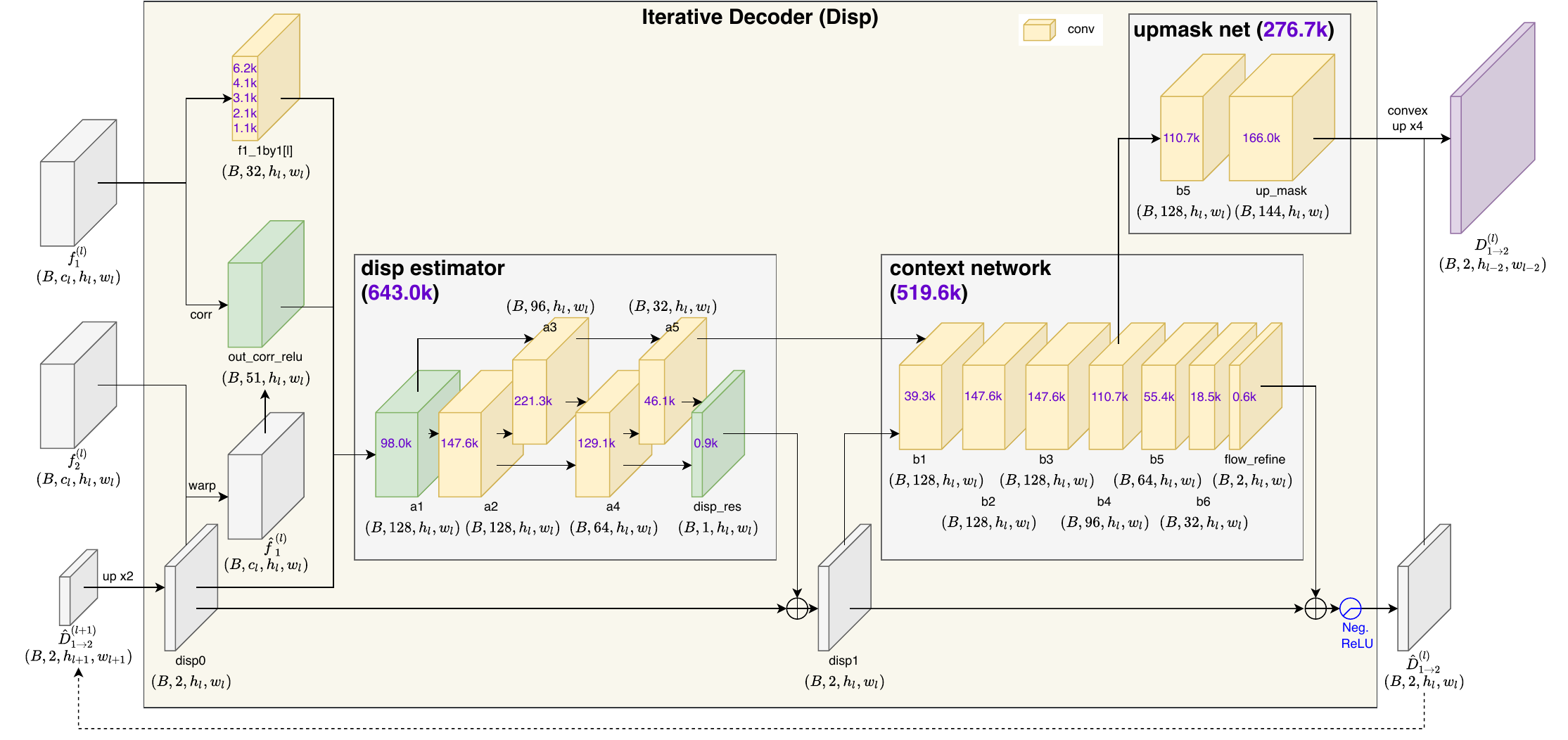}
    \caption{Disp decoder structure (Stage 1)}
    \label{fig:dec_d_app}
 \end{subfigure}
 \caption{Stage-1 model detailed structures}
\end{figure*}

\paragraph{Detailed Structures} The detailed network structure of our encoder, flow decoder, and disparity decoder are shown in \cref{fig:enc_app,fig:dec_f_app,fig:dec_d_app}. The dimension of each intermediate tensor and the number of parameters are noted in the figures.

\paragraph{Redundant Dimension in Disparity}
To reuse modules across flow and disparity decoders, we need to make sure their input sizes to the same module are equal. Therefore, in our implementation, our disparity is estimated as a 2D vector. In the disparity estimator module, which is not shared, the output dimension is set as 1, so we only change the $x$-dimension of the disparity estimate. For other shared modules, we use 2D disparities as both input and output. In the end, we drop the $y$-dimension of our 2D disparity estimate to make sure our final disparity is strictly 1D.

\paragraph{Augmentation as Self-Supervision}
Our network is adapted from ARFlow~\cite{liu2020learning}, which has an ``augmentation as self-supervision'' module. In ARFlow, after each forward pass of the current sample, they apply some random transformations, including appearance transformations (brightness, contrast, hue, saturation, gaussian blur, \etc), spatial transformations (random affine), and occlusion transformations (random cropping), to generate an augmented sample as well as its pseudo-label. Then, they do a second forward pass using the augmented sample and self-supervise the output using the generated pseudo-label. This self-supervision module helps learn flow at occlusion regions and enhance the consistency of optical flow prediction.

We keep their augmentation as self-supervision module in training, and we apply the same transformations to all three input images, $I_{{L1}}, I_{{R1}}, I_{{L2}}$, at the same time to simulate a realistic transformation to the moving stereo camera rig. However, we have to turn off the random rotation part in the random affine transformation to ensure the left and right views are on the same horizontal line. We assume 1D horizontal search for disparities, and if rotations are applied, the true disparity will no longer be 1D, which affects the effectiveness of our disparity decoder.

\subsection{Stage 2}

\begin{figure}[tb]
    \centering
    \includegraphics[width=0.66\linewidth]{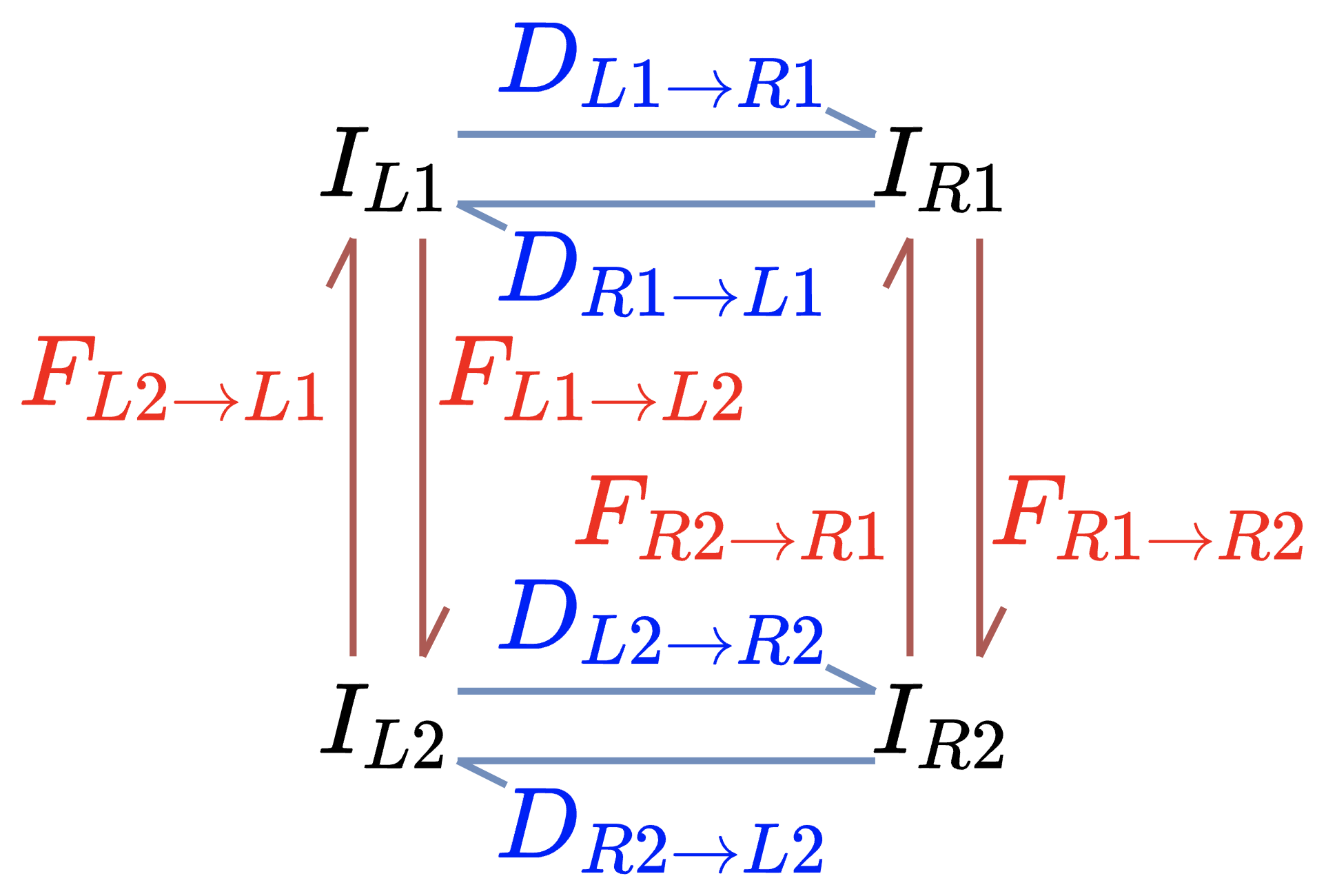}
    \caption{All flows and disparities computed}
    \label{fig:all_fd}
\end{figure}

\paragraph{Computing Occlusion Masks}
We first use the Stage-1 model to generate all the flow and disparities needed for Stage-2 training. In addition to the regular outputs $F_{L1\to L2}, D_{L1\to R1}$ shown in the Stage-1 network, we also compute all other flows and disparities (including backward ones) among the four input images $I_{{L1}}, I_{{R1}}, I_{{L2}}, I_{{R2}}$, as enumerated in \cref{fig:all_fd}. This can be done by switching different inputs to the Stage-1 network. For example, we can compute $F_{L2\to L1}, D_{L2\to R2}$ by feeding $I_{{L2}}, I_{{R2}}, I_{{L1}}$ to the network. Note that if we swap the left and right view, we also need to do a horizontal flip to all input images to ensure the disparity is negative. We can then horizontally flip the disparity output again to recover the original disparity.

With all flows and disparities estimated, we compute all occlusion masks through forward-backward consistency check~\cite{meister2018unflow} as follows. Suppose $F_f$ and $F_b$ are the forward and backward pair of flows (or disparities) and $\bm x$ is a point in the frame. We flag the forward occlusion mask as 1 on $\bm x$ whenever the following constraint holds.
\begin{eqnarray}
 &&  \|F_f(\bm x)  + F_b(\bm x + F_f(\bm x))\|_2^2  \nonumber\\
 &&  < \alpha_1 \left(\|F_f(\bm x)\|_2^2 + \|F_b(\bm x + F_f(\bm x))\|_2^2\right) + \alpha_2, \nonumber   
\end{eqnarray}
where we use hyper-parameters $\alpha_1 =0.01$, $\alpha_2 = 0.5$ as in~\cite{meister2018unflow}. We compute occlusion masks for every forward and backward flow and disparities computed and store them on the disk for later use.

\paragraph{Generating Scene Flow Pseudo-labels for RAFT-3D} The RAFT-3D~\cite{teed2021raft} network also requires a disparity channel input to become RGBD inputs. We can simply use the disparities estimated by the Stage-1 model for that. 

In loss computation, RAFT-3D requires scene flow pseudo-labels, which include the regular 2D optical flow, as well as a disparity change value to indicate the motion on the depth dimension. For the former, we can simply use the previous flow estimate. For the latter, we need to warp previous disparities. For example, to compute scene flow pseudo-label between $I_{{L1}}$ and $I_{{L2}}$, for each point $\bm x$, we warp $D_{L2\to R2}$ and compute the disparity change pseudo-label as follows.
\[
\zeta_{L1\to L2}(\bm x) =D_{L2\to R2}\left(\bm x + F_{L1\to L2}(\bm x)\right) - D_{L1\to R1}(\bm x)
\]

We mask out the unreliable regions of both scene flow estimates and pseudo-labels when we compute loss. Since our flow in Stage 2 is reconstructed from disparity $D_{L1\to R1}$, the unreliable region of the flow estimate is simply the estimated occlusion region of $D_{L1\to R1}$. The unreliable region of pseudo-label is generated as a union of the occlusion regions of $F_{L1\to L2}$, $D_{L1\to R1}$, and the warped $D_{L2\to R2}$.

\paragraph{A Smaller Context Network} Our Stage-2 network is exactly the same as RAFT-3D~\cite{teed2021raft} except for one change on the context network. The original RAFT-3D uses a very large context network copied from FPN~\cite{lin2017feature} so that they can use semantically pre-trained weights to help better distinguish objects. In our experiment, we found empirically that using a much smaller context network such as the same structure of the feature encoder network still gives similar results. Therefore, we stick to the smaller context network to save memory.

\subsection{Stage 3}

\paragraph{Depth Refinement}
Continuing the derivation in the main paper, we need to solve the following over-determined linear system
\[
\left(-\frac{1}{\hat D_1} + \frac{1}{\hat D_1^2}\delta_1\right)\bm\alpha_1 + \left(-\frac{1}{\hat D_2} + \frac{1}{\hat D_2^2}\delta_2\right)\bm\alpha_2 = \bm t,
\]
which can be rewritten as 
\[
\delta_1\frac{\bm \alpha_1}{\hat D_1^2} + \delta_2\frac{\bm \alpha_2}{\hat D_2^2} =\frac{\bm t}{f_xb} + \frac{\bm \alpha_1}{\hat D_1} + \frac{\bm \alpha_2}{\hat D_2}.
\]
Denoting $\bm\beta_1 = \frac{\bm \alpha_1}{\hat D_1^2}$, $\bm\beta_2 = \frac{\bm \alpha_2}{\hat D_2^2}$, $\bm \gamma = \frac{\bm t}{f_xb} + \frac{\bm \alpha_1}{\hat D_1} + \frac{\bm \alpha_2}{\hat D_2}$, we then have
\[
[\bm \beta_1\mid \bm \beta_2]
\left[ \begin{array}{c}
\delta_1 \\
\delta_2      
\end{array}\right]
 = \bm \gamma.
\]
This over-determined linear system can be approximately solved based on least squared error criterion as
\[
\left [
\begin{array}{l}
\delta_1^* \\
\delta_2^*
\end{array}
\right ] = (B^TB)^{-1}B^T\bm \gamma,
\]
where $B = [\bm \beta_1\mid \bm \beta_2] \in \R^{3\times 2}$.

We linearize this equation here instead of solving it in the depth domain (which could be more straightforward) because solving this type of linear systems could be very sensitive in the depth domain. For example, when the 3D motion is along the moving direction of the camera, we have the projection rays from the two consecutive frames almost in parallel, so the $B$ matrix here will be close to rank 1. To avoid this issue, we discard the refinement if $B^TB$ is close to singular (decided by the opencv package automatically).

\section{Experiment Details}

\subsection{Screenshots of Benchmark Results}
\begin{figure}[htb]
    \centering
    \includegraphics[width=\linewidth]{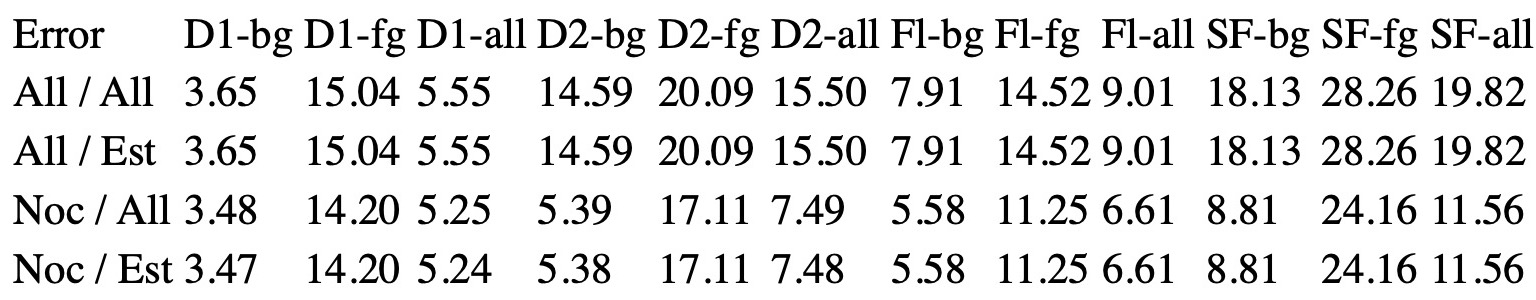}
    \caption{Stage-1 benchmark test result screenshot}
    \label{fig:ss1}
\end{figure}

\begin{figure}[htb]
    \centering
    \includegraphics[width=\linewidth]{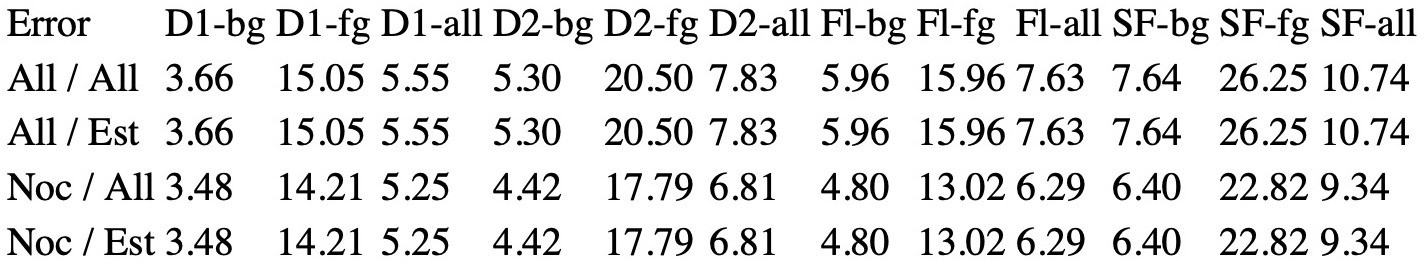}
    \caption{Stage-2 benchmark test result screenshot}
    \label{fig:ss2}
\end{figure}

\begin{figure}[htb]
    \centering
    \includegraphics[width=0.4\linewidth]{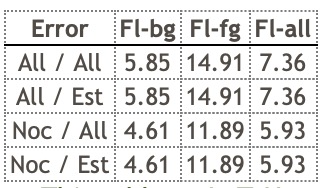}
    \caption{Stage-3 benchmark test result screenshot}
    \label{fig:ss3}
\end{figure}

We show the benchmark test screenshots for all our three stages from the KITTI~\cite{kitti15} website in \cref{fig:ss1,fig:ss2,fig:ss3} to prove that our results are official. Our final-stage result entry (currently anonymous) is also shown on the official website now for your reference.

\subsection{More Qualitative Examples}

\begin{figure*}[htb]
    \centering
    \includegraphics[width=\linewidth]{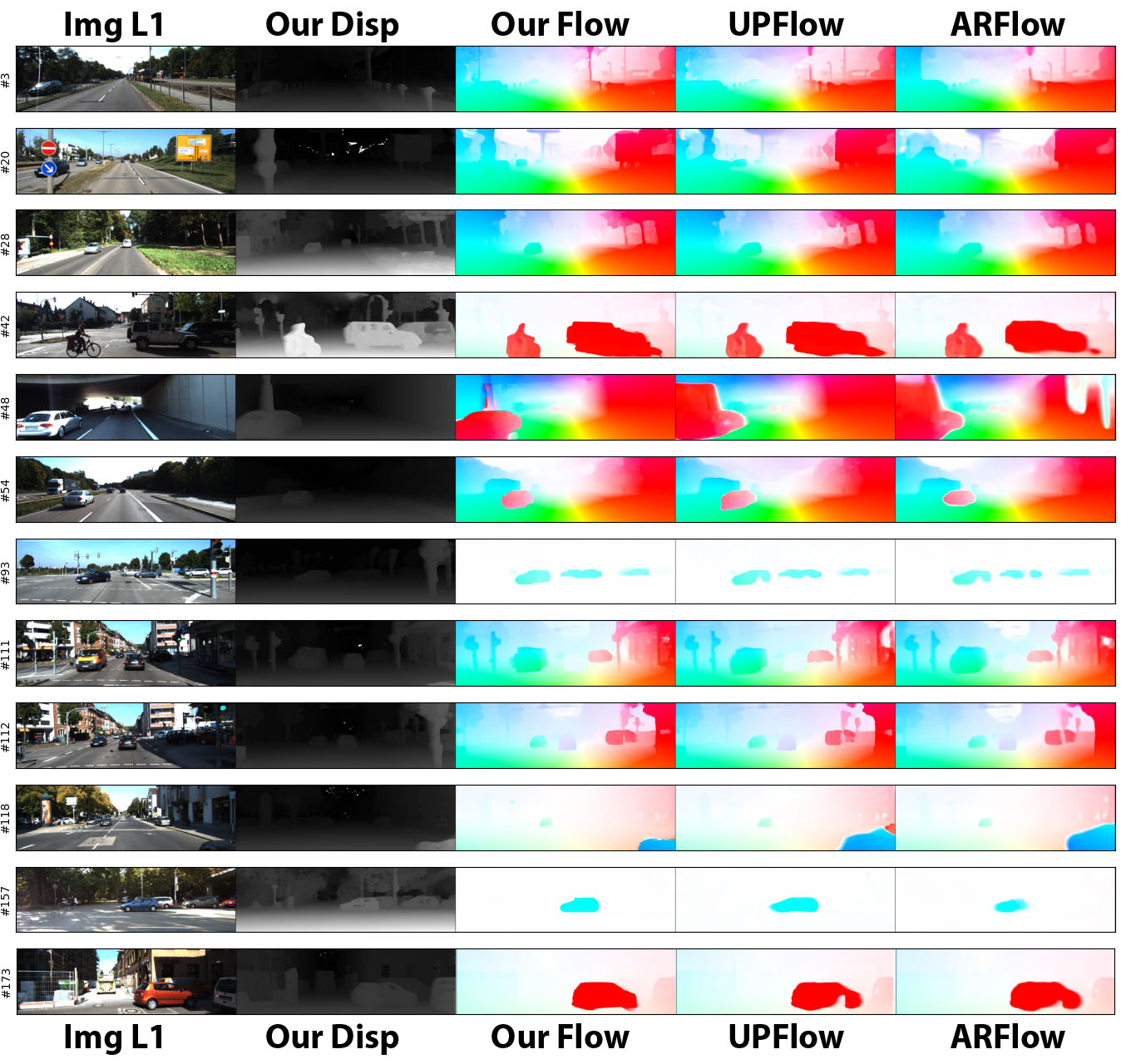}
    \caption{More qualitative examples}
    \label{fig:demo_app}
\end{figure*}

Some more qualitative examples from the KITTI-2015~\cite{kitti15} test set are shown in \cref{fig:demo_app}.


\subsection{Visualizing Rigid Motions Segmentation}

Our rigid motion also implies a soft segmentation of rigid bodies that we can refer to. This can be computed using traditional clustering algorithms based on our 6-DoF map. However, the selection of clustering method also makes a difference here. It may also be helpful if we could have object-level information to regularize our map segmentation. Thus, we leave this for future research.

Alternatively, we can also visualize the map in RGB through PCA. For cases with exactly 3 distinct rigid motions, we can see the segmentation very clearly after normalizing the first three principal component channel-wise to between 0 and 1, as shown in the last figure in the main paper. For most cases that do not have at least three distinct rigid motions, normalizing the top 3 principal components channel-wise may magnify the small changes represented by the second or third color channel. In that case, we suggest normalizing all three channels together. One example of that is shown in our Fig. 1 in the main paper.

\clearpage

\clearpage
{
    \small
    \bibliographystyle{plain}
    \bibliography{main}

\begin{thebibliography}{10}

\bibitem{anton2013elementary}
Howard Anton and Chris Rorres.
\newblock {\em Elementary linear algebra: applications version}.
\newblock John Wiley \& Sons, 2013.

\bibitem{barnes2009patchmatch}
Connelly Barnes, Eli Shechtman, Adam Finkelstein, and Dan~B Goldman.
\newblock Patchmatch: A randomized correspondence algorithm for structural
  image editing.
\newblock {\em ACM TOG}, 28(3):24, 2009.

\bibitem{bastug2017toward}
Ejder Bastug, Mehdi Bennis, Muriel M{\'e}dard, and M{\'e}rouane Debbah.
\newblock Toward interconnected virtual reality: Opportunities, challenges, and
  enablers.
\newblock {\em IEEE Communications Magazine}, 55(6):110--117, 2017.

\bibitem{bayramli2023raft}
Bayram Bayramli, Junhwa Hur, and Hongtao Lu.
\newblock Raft-msf: Self-supervised monocular scene flow using recurrent
  optimizer.
\newblock {\em IJCV}, pages 1--13, 2023.

\bibitem{besl1992method}
Paul~J Besl and Neil~D McKay.
\newblock Method for registration of 3-d shapes.
\newblock In {\em Sensor Fusion IV: Control Paradigms and Data Structures},
  volume 1611, pages 586--606. Spie, 1992.

\bibitem{carion2020end}
Nicolas Carion, Francisco Massa, Gabriel Synnaeve, Nicolas Usunier, Alexander
  Kirillov, and Sergey Zagoruyko.
\newblock End-to-end object detection with transformers.
\newblock In {\em ECCV}, pages 213--229. Springer, 2020.

\bibitem{chang2018pyramid}
Jia-Ren Chang and Yong-Sheng Chen.
\newblock Pyramid stereo matching network.
\newblock In {\em CVPR}, pages 5410--5418, 2018.

\bibitem{chen2020deep}
Guanying Chen, Kai Han, Boxin Shi, Yasuyuki Matsushita, and Kwan-Yee~K Wong.
\newblock Deep photometric stereo for non-lambertian surfaces.
\newblock {\em IEEE TPAMI}, 44(1):129--142, 2020.

\bibitem{cheng2019learning}
Xinjing Cheng, Peng Wang, and Ruigang Yang.
\newblock Learning depth with convolutional spatial propagation network.
\newblock {\em IEEE TPAMI}, 42(10):2361--2379, 2019.

\bibitem{chi2021feature}
Cheng Chi, Qingjie Wang, Tianyu Hao, Peng Guo, and Xin Yang.
\newblock Feature-level collaboration: Joint unsupervised learning of optical
  flow, stereo depth and camera motion.
\newblock In {\em CVPR}, pages 2463--2473, 2021.

\bibitem{dong2022towards}
Xingshuai Dong, Matthew~A Garratt, Sreenatha~G Anavatti, and Hussein~A Abbass.
\newblock Towards real-time monocular depth estimation for robotics: A survey.
\newblock {\em IEEE Trans. Intell. Transport. Sys.}, 23(10):16940--16961, 2022.

\bibitem{dosovitskiy2020image}
Alexey Dosovitskiy, Lucas Beyer, Alexander Kolesnikov, Dirk Weissenborn,
  Xiaohua Zhai, Thomas Unterthiner, Mostafa Dehghani, Matthias Minderer, Georg
  Heigold, Sylvain Gelly, Jakob Uszkoreit, and Neil Houlsby.
\newblock An image is worth 16x16 words: Transformers for image recognition at
  scale.
\newblock In {\em ICLR}, 2021.

\bibitem{dosovitskiy2015flownet}
Alexey Dosovitskiy, Philipp Fischer, Eddy Ilg, Philip Hausser, Caner Hazirbas,
  Vladimir Golkov, Patrick Van Der~Smagt, Daniel Cremers, and Thomas Brox.
\newblock Flownet: Learning optical flow with convolutional networks.
\newblock In {\em ICCV}, pages 2758--2766, 2015.

\bibitem{fischler1981random}
Martin~A Fischler and Robert~C Bolles.
\newblock Random sample consensus: a paradigm for model fitting with
  applications to image analysis and automated cartography.
\newblock {\em ACM Communications}, 24(6):381--395, 1981.

\bibitem{forsyth2002computer}
David~A Forsyth and Jean Ponce.
\newblock {\em Computer vision: a modern approach}.
\newblock prentice hall professional technical reference, 2002.

\bibitem{kitti12}
Andreas Geiger, Philip Lenz, Christoph Stiller, and Raquel Urtasun.
\newblock Vision meets robotics: The kitti dataset.
\newblock {\em IJRR}, 32(11):1231--1237, 2013.

\bibitem{geiger2012we}
Andreas Geiger, Philip Lenz, and Raquel Urtasun.
\newblock Are we ready for autonomous driving? the kitti vision benchmark
  suite.
\newblock In {\em CVPR}, pages 3354--3361. IEEE, 2012.

\bibitem{girshick2015fast}
Ross Girshick.
\newblock Fast r-cnn.
\newblock In {\em ICCV}, pages 1440--1448, 2015.

\bibitem{godard2017unsupervised}
Cl{\'e}ment Godard, Oisin Mac~Aodha, and Gabriel~J Brostow.
\newblock Unsupervised monocular depth estimation with left-right consistency.
\newblock In {\em CVPR}, pages 270--279, 2017.

\bibitem{guo2019group}
Xiaoyang Guo, Kai Yang, Wukui Yang, Xiaogang Wang, and Hongsheng Li.
\newblock Group-wise correlation stereo network.
\newblock In {\em CVPR}, pages 3273--3282, 2019.

\bibitem{hannah1974computer}
Marsha~Jo Hannah.
\newblock {\em Computer matching of areas in stereo images.}
\newblock Stanford University, 1974.

\bibitem{hartley2003multiple}
Richard Hartley and Andrew Zisserman.
\newblock {\em Multiple view geometry in computer vision}.
\newblock Cambridge university press, 2003.

\bibitem{he2016identity}
Kaiming He, Xiangyu Zhang, Shaoqing Ren, and Jian Sun.
\newblock Identity mappings in deep residual networks.
\newblock In {\em ECCV}, pages 630--645. Springer, 2016.

\bibitem{hirschmuller2007stereo}
Heiko Hirschmuller.
\newblock Stereo processing by semiglobal matching and mutual information.
\newblock {\em IEEE TPAMI}, 30(2):328--341, 2007.

\bibitem{ho2020denoising}
Jonathan Ho, Ajay Jain, and Pieter Abbeel.
\newblock Denoising diffusion probabilistic models.
\newblock {\em NeurIPS}, 33:6840--6851, 2020.

\bibitem{horn1981determining}
Berthold~KP Horn and Brian~G Schunck.
\newblock Determining optical flow.
\newblock {\em Artificial Intelligence}, 17(1-3):185--203, 1981.

\bibitem{huang2023self}
Hsin-Ping Huang, Charles Herrmann, Junhwa Hur, Erika Lu, Kyle Sargent, Austin
  Stone, Ming-Hsuan Yang, and Deqing Sun.
\newblock Self-supervised autoflow.
\newblock In {\em CVPR}, pages 11412--11421, 2023.

\bibitem{huang2022flowformer}
Zhaoyang Huang, Xiaoyu Shi, Chao Zhang, Qiang Wang, Ka~Chun Cheung, Hongwei
  Qin, Jifeng Dai, and Hongsheng Li.
\newblock Flowformer: A transformer architecture for optical flow.
\newblock In {\em ECCV}, pages 668--685. Springer, 2022.

\bibitem{hur2019iterative}
Junhwa Hur and Stefan Roth.
\newblock Iterative residual refinement for joint optical flow and occlusion
  estimation.
\newblock In {\em CVPR}, pages 5754--5763, 2019.

\bibitem{hur2020self}
Junhwa Hur and Stefan Roth.
\newblock Self-supervised monocular scene flow estimation.
\newblock In {\em CVPR}, pages 7396--7405, 2020.

\bibitem{janai2018unsupervised}
Joel Janai, Fatma Guney, Anurag Ranjan, Michael Black, and Andreas Geiger.
\newblock Unsupervised learning of multi-frame optical flow with occlusions.
\newblock In {\em ECCV}, pages 690--706, 2018.

\bibitem{jiang2021learning}
Shihao Jiang, Dylan Campbell, Yao Lu, Hongdong Li, and Richard Hartley.
\newblock Learning to estimate hidden motions with global motion aggregation.
\newblock In {\em ICCV}, pages 9772--9781, 2021.

\bibitem{jiao2021effiscene}
Yang Jiao, Trac~D Tran, and Guangming Shi.
\newblock Effiscene: Efficient per-pixel rigidity inference for unsupervised
  joint learning of optical flow, depth, camera pose and motion segmentation.
\newblock In {\em CVPR}, pages 5538--5547, 2021.

\bibitem{joung2019unsupervised}
Sunghun Joung, Seungryong Kim, Kihong Park, and Kwanghoon Sohn.
\newblock Unsupervised stereo matching using confidential correspondence
  consistency.
\newblock {\em IEEE Trans. Intell. Transport. Sys.}, 21(5):2190--2203, 2019.

\bibitem{jung2023anyflow}
Hyunyoung Jung, Zhuo Hui, Lei Luo, Haitao Yang, Feng Liu, Sungjoo Yoo, Rakesh
  Ranjan, and Denis Demandolx.
\newblock Anyflow: Arbitrary scale optical flow with implicit neural
  representation.
\newblock In {\em CVPR}, pages 5455--5465, 2023.

\bibitem{kendall2017end}
Alex Kendall, Hayk Martirosyan, Saumitro Dasgupta, Peter Henry, Ryan Kennedy,
  Abraham Bachrach, and Adam Bry.
\newblock End-to-end learning of geometry and context for deep stereo
  regression.
\newblock In {\em ICCV}, pages 66--75, 2017.

\bibitem{kim2022cross}
Hannah~Halin Kim, Shuzhi Yu, Shuai Yuan, and Carlo Tomasi.
\newblock Cross-attention transformer for video interpolation.
\newblock In {\em ACCVW}, pages 320--337, 2022.

\bibitem{kingma2014adam}
Diederik~P Kingma and Jimmy Ba.
\newblock Adam: A method for stochastic optimization.
\newblock {\em ICLR}, 2014.

\bibitem{kolmogorov2005convergent}
Vladimir Kolmogorov.
\newblock Convergent tree-reweighted message passing for energy minimization.
\newblock In {\em International Workshop on Artificial Intelligence and
  Statistics}, pages 182--189. PMLR, 2005.

\bibitem{kong2023mitigating}
Fanjie Kong, Shuai Yuan, Weituo Hao, and Ricardo Henao.
\newblock Mitigating test-time bias for fair image retrieval.
\newblock In {\em NeurIPS}, 2023.

\bibitem{krizhevsky2012imagenet}
Alex Krizhevsky, Ilya Sutskever, and Geoffrey~E Hinton.
\newblock Imagenet classification with deep convolutional neural networks.
\newblock {\em NeurIPS}, 25, 2012.

\bibitem{lai2019bridging}
Hsueh-Ying Lai, Yi-Hsuan Tsai, and Wei-Chen Chiu.
\newblock Bridging stereo matching and optical flow via spatiotemporal
  correspondence.
\newblock In {\em CVPR}, pages 1890--1899, 2019.

\bibitem{lepetit2009ep}
Vincent Lepetit, Francesc Moreno-Noguer, and Pascal Fua.
\newblock Epnp: An accurate o(n) solution to the pnp problem.
\newblock {\em IJCV}, 81:155--166, 2009.

\bibitem{li2021unsupervised}
Ang Li, Zejian Yuan, Yonggen Ling, Wanchao Chi, Shenghao Zhang, and Chong
  Zhang.
\newblock Unsupervised occlusion-aware stereo matching with directed disparity
  smoothing.
\newblock {\em IEEE Trans. Intell. Transport. Sys.}, 23(7):7457--7468, 2021.

\bibitem{lin2017feature}
Tsung-Yi Lin, Piotr Doll{\'a}r, Ross Girshick, Kaiming He, Bharath Hariharan,
  and Serge Belongie.
\newblock Feature pyramid networks for object detection.
\newblock In {\em CVPR}, pages 2117--2125, 2017.

\bibitem{lipson2021raft}
Lahav Lipson, Zachary Teed, and Jia Deng.
\newblock Raft-stereo: Multilevel recurrent field transforms for stereo
  matching.
\newblock In {\em 3DV}, pages 218--227. IEEE, 2021.

\bibitem{liu2019unsupervised}
Liang Liu, Guangyao Zhai, Wenlong Ye, and Yong Liu.
\newblock Unsupervised learning of scene flow estimation fusing with local
  rigidity.
\newblock In {\em IJCAI}, pages 876--882, 2019.

\bibitem{liu2020learning}
Liang Liu, Jiangning Zhang, Ruifei He, Yong Liu, Yabiao Wang, Ying Tai, Donghao
  Luo, Chengjie Wang, Jilin Li, and Feiyue Huang.
\newblock Learning by analogy: Reliable supervision from transformations for
  unsupervised optical flow estimation.
\newblock In {\em CVPR}, pages 6489--6498, 2020.

\bibitem{liu2019ddflow}
Pengpeng Liu, Irwin King, Michael~R Lyu, and Jia Xu.
\newblock Ddflow: Learning optical flow with unlabeled data distillation.
\newblock In {\em AAAI}, volume~33, pages 8770--8777, 2019.

\bibitem{liu2020flow2stereo}
Pengpeng Liu, Irwin King, Michael~R Lyu, and Jia Xu.
\newblock Flow2stereo: Effective self-supervised learning of optical flow and
  stereo matching.
\newblock In {\em CVPR}, pages 6648--6657, 2020.

\bibitem{liu2019selflow}
Pengpeng Liu, Michael Lyu, Irwin King, and Jia Xu.
\newblock Selflow: Self-supervised learning of optical flow.
\newblock In {\em CVPR}, pages 4571--4580, 2019.

\bibitem{lucas1981iterative}
Bruce~D Lucas and Takeo Kanade.
\newblock An iterative image registration technique with an application to
  stereo vision.
\newblock In {\em IJCAI}, volume~2, pages 674--679, 1981.

\bibitem{luo2019every}
Chenxu Luo, Zhenheng Yang, Peng Wang, Yang Wang, Wei Xu, Ram Nevatia, and Alan
  Yuille.
\newblock Every pixel counts++: Joint learning of geometry and motion with 3d
  holistic understanding.
\newblock {\em IEEE TPAMI}, 42(10):2624--2641, 2019.

\bibitem{luo2021upflow}
Kunming Luo, Chuan Wang, Shuaicheng Liu, Haoqiang Fan, Jue Wang, and Jian Sun.
\newblock Upflow: Upsampling pyramid for unsupervised optical flow learning.
\newblock In {\em CVPR}, pages 1045--1054, 2021.

\bibitem{meister2018unflow}
Simon Meister, Junhwa Hur, and Stefan Roth.
\newblock Unflow: Unsupervised learning of optical flow with a bidirectional
  census loss.
\newblock In {\em AAAI}, volume~32, 2018.

\bibitem{kitti15}
Moritz Menze and Andreas Geiger.
\newblock Object scene flow for autonomous vehicles.
\newblock In {\em CVPR}, 2015.

\bibitem{mohamed2014illumination}
Mahmoud~A Mohamed, Hatem~A Rashwan, B{\"a}rbel Mertsching, Miguel~Angel
  Garc{\'\i}a, and Domenec Puig.
\newblock Illumination-robust optical flow using a local directional pattern.
\newblock {\em IEEE TCSVT}, 24(9):1499--1508, 2014.

\bibitem{NEURIPS2019_9015}
Adam Paszke, Sam Gross, Francisco Massa, Adam Lerer, James Bradbury, Gregory
  Chanan, Trevor Killeen, Zeming Lin, Natalia Gimelshein, Luca Antiga, Alban
  Desmaison, Andreas Kopf, Edward Yang, Zachary DeVito, Martin Raison, Alykhan
  Tejani, Sasank Chilamkurthy, Benoit Steiner, Lu~Fang, Junjie Bai, and Soumith
  Chintala.
\newblock Pytorch: An imperative style, high-performance deep learning library.
\newblock In {\em NeurIPS}, pages 8024--8035. Curran Associates, Inc., 2019.

\bibitem{pearson1901liii}
Karl Pearson.
\newblock Liii. on lines and planes of closest fit to systems of points in
  space.
\newblock {\em The London, Edinburgh, and Dublin Philosophical Magazine and
  Journal of Science}, 2(11):559--572, 1901.

\bibitem{ranjan2019competitive}
Anurag Ranjan, Varun Jampani, Lukas Balles, Kihwan Kim, Deqing Sun, Jonas
  Wulff, and Michael~J Black.
\newblock Competitive collaboration: Joint unsupervised learning of depth,
  camera motion, optical flow and motion segmentation.
\newblock In {\em CVPR}, pages 12240--12249, 2019.

\bibitem{ren2017unsupervised}
Zhe Ren, Junchi Yan, Bingbing Ni, Bin Liu, Xiaokang Yang, and Hongyuan Zha.
\newblock Unsupervised deep learning for optical flow estimation.
\newblock In {\em AAAI}, 2017.

\bibitem{ren2019fusion}
Zhile Ren, Orazio Gallo, Deqing Sun, Ming-Hsuan Yang, Erik~B Sudderth, and Jan
  Kautz.
\newblock A fusion approach for multi-frame optical flow estimation.
\newblock In {\em WACV}, pages 2077--2086. IEEE, 2019.

\bibitem{revaud2015epicflow}
Jerome Revaud, Philippe Weinzaepfel, Zaid Harchaoui, and Cordelia Schmid.
\newblock Epicflow: Edge-preserving interpolation of correspondences for
  optical flow.
\newblock In {\em CVPR}, pages 1164--1172, 2015.

\bibitem{ruder2017overview}
Sebastian Ruder.
\newblock An overview of multi-task learning in deep neural networks.
\newblock {\em arXiv preprint arXiv:1706.05098}, 2017.

\bibitem{sawhney19943d}
Sawhney.
\newblock 3d geometry from planar parallax.
\newblock In {\em CVPR}, pages 929--934. IEEE, 1994.

\bibitem{schonberger2016structure}
Johannes~L Schonberger and Jan-Michael Frahm.
\newblock Structure-from-motion revisited.
\newblock In {\em CVPR}, pages 4104--4113, 2016.

\bibitem{shi1994good}
Jianbo Shi et~al.
\newblock Good features to track.
\newblock In {\em CVPR}, pages 593--600. IEEE, 1994.

\bibitem{shi2023flowformer++}
Xiaoyu Shi, Zhaoyang Huang, Dasong Li, Manyuan Zhang, Ka~Chun Cheung, Simon
  See, Hongwei Qin, Jifeng Dai, and Hongsheng Li.
\newblock Flowformer++: Masked cost volume autoencoding for pretraining optical
  flow estimation.
\newblock In {\em CVPR}, pages 1599--1610, 2023.

\bibitem{smith2019super}
Leslie~N Smith and Nicholay Topin.
\newblock Super-convergence: Very fast training of neural networks using large
  learning rates.
\newblock In {\em Artificial Intelligence and Machine Learning for Multi-Domain
  Operations Applications}, volume 11006, pages 369--386. SPIE, 2019.

\bibitem{stone2021smurf}
Austin Stone, Daniel Maurer, Alper Ayvaci, Anelia Angelova, and Rico
  Jonschkowski.
\newblock Smurf: Self-teaching multi-frame unsupervised raft with full-image
  warping.
\newblock In {\em CVPR}, pages 3887--3896, 2021.

\bibitem{sun2018pwc}
Deqing Sun, Xiaodong Yang, Ming-Yu Liu, and Jan Kautz.
\newblock Pwc-net: Cnns for optical flow using pyramid, warping, and cost
  volume.
\newblock In {\em CVPR}, pages 8934--8943, 2018.

\bibitem{teed2020raft}
Zachary Teed and Jia Deng.
\newblock Raft: Recurrent all-pairs field transforms for optical flow.
\newblock In {\em ECCV}, pages 402--419. Springer, 2020.

\bibitem{teed2021raft}
Zachary Teed and Jia Deng.
\newblock Raft-3d: Scene flow using rigid-motion embeddings.
\newblock In {\em CVPR}, pages 8375--8384, 2021.

\bibitem{teed2021tangent}
Zachary Teed and Jia Deng.
\newblock Tangent space backpropagation for 3d transformation groups.
\newblock In {\em Proceedings of the IEEE/CVF Conference on Computer Vision and
  Pattern Recognition (CVPR)}, 2021.

\bibitem{thurston1997three}
William P~HG Thurston.
\newblock {\em Three-Dimensional Geometry and Topology, Volume 1: Volume 1}.
\newblock Princeton university press, 1997.

\bibitem{torr1998geometric}
Philip~HS Torr.
\newblock Geometric motion segmentation and model selection.
\newblock {\em Philosophical Transactions of the Royal Society of London.
  Series A: Mathematical, Physical and Engineering Sciences},
  356(1740):1321--1340, 1998.

\bibitem{torr1999problem}
Philip~HS Torr, Andrew~W Fitzgibbon, and Andrew Zisserman.
\newblock The problem of degeneracy in structure and motion recovery from
  uncalibrated image sequences.
\newblock {\em IJCV}, 32:27--44, 1999.

\bibitem{ullman1979interpretation}
Shimon Ullman.
\newblock The interpretation of structure from motion.
\newblock {\em Proceedings of the Royal Society of London. Series B. Biological
  Sciences}, 203(1153):405--426, 1979.

\bibitem{vidal2006two}
Ren{\'e} Vidal, Yi~Ma, Stefano Soatto, and Shankar Sastry.
\newblock Two-view multibody structure from motion.
\newblock {\em IJCV}, 68(1):7--25, 2006.

\bibitem{vidal2003optimal}
Ren{\'e} Vidal and Shankar Sastry.
\newblock Optimal segmentation of dynamic scenes from two perspective views.
\newblock In {\em CVPR}, volume~2, pages II--II. IEEE, 2003.

\bibitem{vijayanarasimhan2017sfm}
Sudheendra Vijayanarasimhan, Susanna Ricco, Cordelia Schmid, Rahul Sukthankar,
  and Katerina Fragkiadaki.
\newblock Sfm-net: Learning of structure and motion from video.
\newblock {\em arXiv preprint arXiv:1704.07804}, 2017.

\bibitem{wang2020unsupervised}
Guangming Wang, Chi Zhang, Hesheng Wang, Jingchuan Wang, Yong Wang, and Xinlei
  Wang.
\newblock Unsupervised learning of depth, optical flow and pose with occlusion
  from 3d geometry.
\newblock {\em IEEE Trans. Intell. Transport. Sys.}, 23(1):308--320, 2020.

\bibitem{wang2019unos}
Yang Wang, Peng Wang, Zhenheng Yang, Chenxu Luo, Yi~Yang, and Wei Xu.
\newblock Unos: Unified unsupervised optical-flow and stereo-depth estimation
  by watching videos.
\newblock In {\em CVPR}, pages 8071--8081, 2019.

\bibitem{wang2018occlusion}
Yang Wang, Yi~Yang, Zhenheng Yang, Liang Zhao, Peng Wang, and Wei Xu.
\newblock Occlusion aware unsupervised learning of optical flow.
\newblock In {\em CVPR}, pages 4884--4893, 2018.

\bibitem{xu2023unifying}
Haofei Xu, Jing Zhang, Jianfei Cai, Hamid Rezatofighi, Fisher Yu, Dacheng Tao,
  and Andreas Geiger.
\newblock Unifying flow, stereo and depth estimation.
\newblock {\em IEEE TPAMI}, 2023.

\bibitem{xu20193d}
Xun Xu, Loong-Fah Cheong, and Zhuwen Li.
\newblock 3d rigid motion segmentation with mixed and unknown number of models.
\newblock {\em IEEE TPAMI}, 43(1):1--16, 2019.

\bibitem{yang2021learning}
Gengshan Yang and Deva Ramanan.
\newblock Learning to segment rigid motions from two frames.
\newblock In {\em CVPR}, pages 1266--1275, 2021.

\bibitem{yin2018geonet}
Zhichao Yin and Jianping Shi.
\newblock Geonet: Unsupervised learning of dense depth, optical flow and camera
  pose.
\newblock In {\em CVPR}, pages 1983--1992, 2018.

\bibitem{yu2016back}
Jason~J. Yu, Adam~W. Harley, and Konstantinos~G. Derpanis.
\newblock Back to basics: Unsupervised learning of optical flow via brightness
  constancy and motion smoothness.
\newblock In Gang Hua and Herv{\'e} J{\'e}gou, editors, {\em ECCVW}, pages
  3--10, Cham, 2016. Springer International Publishing.

\bibitem{yu2022unsupervised}
Shuzhi Yu, Hannah~H Kim, Shuai Yuan, and Carlo Tomasi.
\newblock Unsupervised flow refinement near motion boundaries.
\newblock In {\em BMVC}. {BMVA} Press, 2022.

\bibitem{yuan2007detecting}
Chang Yuan, Gerard Medioni, Jinman Kang, and Isaac Cohen.
\newblock Detecting motion regions in the presence of a strong parallax from a
  moving camera by multiview geometric constraints.
\newblock {\em IEEE TPAMI}, 29(9):1627--1641, 2007.

\bibitem{yuan2022optical}
Shuai Yuan, Xian Sun, Hannah Kim, Shuzhi Yu, and Carlo Tomasi.
\newblock Optical flow training under limited label budget via active learning.
\newblock In {\em ECCV}, pages 410--427, 2022.

\bibitem{yuan2023semarflow}
Shuai Yuan, Shuzhi Yu, Hannah Kim, and Carlo Tomasi.
\newblock Semarflow: Injecting semantics into unsupervised optical flow
  estimation for autonomous driving.
\newblock In {\em ICCV}, pages 9566--9577, October 2023.

\bibitem{zabih1994non}
Ramin Zabih and John Woodfill.
\newblock Non-parametric local transforms for computing visual correspondence.
\newblock In {\em ECCV}, pages 151--158. Springer, 1994.

\bibitem{zhang2019ga}
Feihu Zhang, Victor Prisacariu, Ruigang Yang, and Philip~HS Torr.
\newblock Ga-net: Guided aggregation net for end-to-end stereo matching.
\newblock In {\em CVPR}, pages 185--194, 2019.

\bibitem{zhang2021separable}
Feihu Zhang, Oliver~J Woodford, Victor~Adrian Prisacariu, and Philip~HS Torr.
\newblock Separable flow: Learning motion cost volumes for optical flow
  estimation.
\newblock In {\em ICCV}, pages 10807--10817, 2021.

\bibitem{zhou2017unsupervised_st}
Chao Zhou, Hong Zhang, Xiaoyong Shen, and Jiaya Jia.
\newblock Unsupervised learning of stereo matching.
\newblock In {\em ICCV}, pages 1567--1575, 2017.

\bibitem{zhou2017unsupervised}
Tinghui Zhou, Matthew Brown, Noah Snavely, and David~G Lowe.
\newblock Unsupervised learning of depth and ego-motion from video.
\newblock In {\em CVPR}, pages 1851--1858, 2017.

\bibitem{zou2018df}
Yuliang Zou, Zelun Luo, and Jia-Bin Huang.
\newblock Df-net: Unsupervised joint learning of depth and flow using
  cross-task consistency.
\newblock In {\em ECCV}, pages 36--53, 2018.

\end{thebibliography}
}

\end{document}